\documentclass{article}

\usepackage{microtype}
\usepackage{graphicx}
\usepackage{subfigure}
\usepackage{booktabs} 
\usepackage{fancyvrb}

\usepackage{amsmath,amsfonts,bm,amsthm}

\usepackage[boxed]{algorithm}
\usepackage[noend]{algpseudocode}
\algrenewcommand\alglinenumber[1]{\scriptsize #1}

\def \bx{\boldsymbol{x}}

\def \by{\boldsymbol{y}}
\def \bz{\boldsymbol{z}}

\def \bv{\boldsymbol{v}}

\def \bh{\boldsymbol{h}}
\def \bb{\boldsymbol{b}}

\def \bu{\boldsymbol{u}}
\def \bv{\boldsymbol{v}}

\def \bw{\boldsymbol{w}}
\def \be{\boldsymbol{e}}

\def \bA{\boldsymbol{A}}

\def \bB{\boldsymbol{B}}
\def \bV{\boldsymbol{V}}
\def \bU{\boldsymbol{U}}
\def \bC{\boldsymbol{C}}
\def \bZ{\boldsymbol{Z}}

\def \bW{\boldsymbol{W}}

\def \bM{\boldsymbol{M}}

\def \bQ{\boldsymbol{Q}}

\def \bR{\boldsymbol{R}}

\def \bSigma{\boldsymbol{\Sigma}}

\usepackage{comment}
\usepackage{soul}
\usepackage{bbm}

\usepackage{enumitem}

\newtheorem{thm}{Theorem}

\newtheorem{prop}{Proposition}

\newtheorem{defn}{Definition}

\def\J{{\boldsymbol{J}}}
\def\R{{\mathbb R}}

\def\Indic{\mathbbm{1}}

\newcommand{\qq}{\vspace*{-2mm}}

\newcommand\bdot[1]{\overset{\tiny\bullet}{#1}{}}

\usepackage{tikz}
\usetikzlibrary{fit,positioning}
\tikzstyle{block} = [rectangle, draw, fill=blue!20, 
    text width=12.8em, text centered, rounded corners, minimum height=4em]
\tikzstyle{line} = [draw, -latex']
\tikzstyle{cloud} = [draw, ellipse,fill=red!20, node distance=3cm,
    minimum height=2em]
\usetikzlibrary{calc}
\tikzset{
  plate/.style={draw, shape=rectangle, rounded corners=0.5ex, thick,
    minimum width=3.1cm, text width=3.1cm, align=right, inner sep=10pt, inner ysep=8pt,label={[xshift=-28pt,yshift=14pt]south east:#1}}
}
\tikzset{
  plate2/.style={draw, shape=rectangle, rounded corners=0.5ex, thick,
    minimum width=1.1cm, text width=1.1cm, align=right, inner sep=10pt, inner ysep=8pt,label={[xshift=-43pt,yshift=14pt]south east:#1}}
}
\tikzset{
  plate3/.style={draw, shape=rectangle, rounded corners=0.5ex, thick,
    minimum width=1.1cm, text width=1.1cm, align=right, inner sep=10pt, inner ysep=8pt,label={[xshift=-20pt,yshift=14pt]south east:#1}}
}
\usepackage{amssymb}
\usepackage{hyperref}
\usepackage{url}
\usepackage{makecell}
\usepackage{booktabs}
\usepackage{listings}
\usepackage{highlighit}
\usepackage{hyperref}

\usepackage[accepted]{icml2021}

\icmltitlerunning{Fast Jacobian-Vector Product for Deep Networks}
\usepackage{multirow}

\begin{document}

\twocolumn[
\icmltitle{Fast Jacobian-Vector Product for Deep Networks}



\icmlsetsymbol{equal}{*}
\begin{icmlauthorlist}
\icmlauthor{Randall~Balestriero}{to}
\icmlauthor{Richard~G.~Baraniuk}{to}
\end{icmlauthorlist}

\icmlaffiliation{to}{Department of Electrical and Computer Engineering, Rice University, Houston, TX}
\icmlcorrespondingauthor{Randall Balestriero}{randallbalestriero@gmail.com}
\icmlkeywords{Deep Learning, Piecewise Linear Networks, Splines, Jacobian Vector Products, forward-mode differentiation, automatic differentiation}

\vskip 0.3in
]



\printAffiliationsAndNotice{\icmlEqualContribution} 

\begin{abstract}
Jacobian-vector products (JVPs) form the backbone of many recent developments in  Deep Networks (DNs), with applications including faster constrained optimization, regularization with generalization guarantees, and adversarial example sensitivity assessments. 
Unfortunately, JVPs are computationally expensive for real world DN architectures and require the use of automatic differentiation to avoid manually adapting the JVP program when changing the DN architecture. 
We propose a novel method to quickly compute JVPs for any DN that employ Continuous Piecewise Affine (e.g., leaky-ReLU, max-pooling, maxout, etc.) nonlinearities. 
We show that our technique is on average $2\times$ faster than the fastest alternative over $13$ DN architectures and across various hardware.
In addition, our solution does not require automatic differentiation and is thus easy to deploy in software, requiring only the modification of a few lines of codes that do not depend on the DN architecture.
\end{abstract}


\section{Introduction}

Deep (Neural) Networks (DNs) are used throughout the spectrum of machine learning tasks ranging from image classification \citep{lawrence1997face,lin2013network,zagoruyko2016wide} to financial fraud prediction \citep{rushin2017horse,roy2018deep,zheng2018generative}. The ability of DNs to reach super-human performances added great momentum into the deployment of those techniques into real life applications e.g. self-driving cars \citep{rao2018deep}, fire prediction \citep{kim2019video} or seismic activity prediction \citep{seydoux2020clustering}. The main ingredients behind this success roughly fall into three camps: (i) specialized optimization algorithms such as Adam \citep{kingma2014adam} coupled with carefully designed learning rate schedules \citep{smith2017cyclical} and/or mini-batch size \citep{devarakonda2017adabatch,wang2017closer}, (ii) novel regularization techniques which can be explicit as with orthogonal/sparse weights penalties \citep{cohen2016group,8877742,wang2020orthogonal} or implicit as with dropout \citep{molchanov2017variational} or data-augmentation \citep{taylor2017improving,perez2017effectiveness,hernandez2018data,shorten2019survey}; and (iii) adaptivity of the DN architecture to the data and task at hand thanks to expert knowledge \citep{graves2013hybrid,seydoux2020clustering} or automated Neural Architecture Search \citep{elsken2019neural}.

Despite DNs' astonishing performance, their practical deployment in the real-world remains hazardous, requiring tremendous tweaking and the implementation of various safe guard measures \citep{marcus2018deep} to avoid failure cases that can lead to injuries or death, but also to ensure fairness and unbiased behaviors \citep{gianfrancesco2018potential}. One path for overcoming those challenges in a more principled way is to develop novel, specialized methods for each part of the deep learning pipeline (optimization/regularization/architecture).

Recent techniques aiming at tackling those shortcomings have been proposed. Most of those techniques rely on {\em Jacobian-vector products} (JVPs) \citep{hirsch1974differential}, which are defined as the projection of a given vector onto the Jacobian matrix of an operator, i.e. the DN in our case. The JVP captures crucial information on the local geometry of the DN input-output mapping which is one of the main reason behind its popularity. For example, \citet{dehaene2020iterative} leverage the JVP to better control the denoising/data generation process of autoencoders via explicit manifold projections (generated samples are repeatedly projected onto the autoencoder Jacobian matrix); \citet{9296823} assess the sensitivity of DNs to adversarial examples (noise realizations are projected onto the DN Jacobian matrix); \citet{cosentino2020provable} guarantees the generalization performance of DNs based on a novel JVP based measure (neighboring data samples are projected onto the DN Jacobian matrix and compared); \citet{marquez2017imposing} trains DNs with explicit input-output constraints (the constraints are cast into JVPs and evaluated as part of the DN training procedure). Beyond those recent developments, JVPs have a long history of being employed to probe models and get precise understandings of the models' properties e.g. input sensitivity, intrinsic dimension, conditioning \citep{gudmundsson1995small,bentbib2015block,balestriero2020max}.
\textit{To enable the deployment of these techniques in practical scenarios, it is crucial to evaluate JVPs as fast as possible and without relying on automatic differentiation, a feature primarily implemented in development libraries but often missing in software deployed in production.}

In this paper, we propose a novel method to evaluate JVPs that is inspired by recent theoretical studies of DNs employing Continuous Piecewise Affine (CPA) nonlinearities and by Forward-mode Differentiation (FD). In short, the CPA property allows to specialize FD to be evaluate only by feeding two inputs to a DN while holding the DN internal state fixed, and combining the two produced outputs such that the final result corresponds to the desired JVP. Our solution offers the following benefits:
\begin{itemize}[noitemsep,nolistsep,topsep=0pt,leftmargin=10pt]
    \item {\bf Speed}: $2\times$ faster on average over the fastest alternative solution with empirical validation on $13$ different DN architectures (including Recurrent Neural Networks, ResNets, DenseNets, VGGs, UNets) and with speed-ups consistent across hardware.
    
    \item {\bf Simple implementation}: Implementing our solution for any DN requires to change only a few lines of code in total, regardless of the architecture being considered. This allows our method to be employed inside the architecture search loop if needed.
    
    \item {\bf Does not employ automatic differentiation}: Our solution can be implemented in any software/library including the ones that are not equipped with automatic differentiation, allowing deployment of the method on production specific hardware/software.
\end{itemize}
While our method is most simple to implement for CPA DNs (a constraint embracing most of the current state-of-the-art architectures), we also demonstrate how to employ it for smooth DN's layers such as batch-normalization, which is CPA only during testing.

This paper is organized as follows: Sec.~\ref{sec:background} reviews current JVPs solutions and thoroughly presents notations from CPA DNs, Sec.~\ref{sec:fastJVP} presents our proposed solution where detailed implementation is provided in Sec.~\ref{sec:implementation} and careful empirical validations are performed in Sec.~\ref{sec:validation}; Sec.~\ref{sec:extension} presents further results on how JVPs are crucial to evaluate many insightful statistics of CPA DNs such as the largest eigenvalues of the DN Jacobian matrix.

\section{Jacobian-Vector Products and Continuous Piecewise Affine Deep Networks}
\label{sec:background}

{\bf Jacobian-Vector Products (JVPs).}~The  JVP  operation, also called Rop (short for right-operation), is abbreviated as ${\rm Rop}(f,\bx,v)$ and computes the directional derivative, with direction $\bu\in \mathbb{R}^D$, of the multi-dimensional operator $f:\mathbb{R}^{D}\mapsto \mathbb{R}^{K}$ with respect to the input $\bx \in \mathbb{R}^D$ as
\begin{equation}
    {\rm Rop}(f,\bx,\bu)\triangleq \J_{f}(\bx)\bu.\label{eq:rop}
\end{equation}
Throughout our study, we will assume that $f$ is differentiable at $\bx$, $\J_{f}(\bx)$ denotes the Jacobian of $f$ evaluated at $\bx$. The ${\rm Rop}$ operator defines a linear operator from $\mathbb{R}^D$ to $\mathbb{R}^K$ with respect to the $\bu$ argument, additional details can be found in Appendix~\ref{sec:detailsJVP}; throughout this paper we will use the terms JVP and Rop interchangeably.
The vector-Jacobian product (VJP) operation, also called Lop, is abbreviated as ${\rm Lop}(f,\bx,\bv)$ with direction $\bv \in \mathbb{R}^K$ and computes the adjoint directional derivative
\begin{equation}
    {\rm Lop}(f,\bx,\bv)\triangleq \bv^T(\J_{f}(\bx)).
\end{equation}
For a thorough review of those operations, we refer the reader to \citet{paszke2017automatic,baydin2017automatic,elliott2018simple}; throughout this paper we will use the terms VJP and Lop interchangeably.

In practice, evaluating the Rop given a differentiable operator $f$ can be done naturally through forward-mode automatic differentiation which builds ${\rm Rop}(f,\bx,\bu)$ while forwarding $\bx$ through the computational graph of $f$. However, it can also be done by employing the backward-mode automatic differentiation \citep{linnainmaa1976taylor} i.e. traversing the computational graph of $f$ from output to input, via multiple Lop calls via
\begin{align}
    {\rm Rop}(f,\bx,\bu)=\J_{f}(\bx)\bu=\begin{pmatrix}{\rm Lop}(f,\bx,\be_1)\\ \vdots \\ {\rm Lop}(f,w,\be_K)\end{pmatrix}\bu,\label{eq:JVPtimes}
\end{align}
where $K$ is the length of the output vector of the mapping $f$. Practical implementations of the above often resort to parallel computations of each $Lop$ forming the rows of the Jacobian matrix reducing computation time at the cost of greater memory requirements. Alternatively, an interesting computational trick recently brought to light in \citet{trick} allows to compute the $Rop$ from two $Lop$ calls only (instead of K) as follows
\begin{align}
    {\rm Rop}(f,\bx,\bu)={\rm Lop}({\rm Lop}(f,\bx,\bv)^T,\bv,\bu)^T\label{eq:doubleVJP},
\end{align}
detailed derivation can be found in Appendix~\ref{sec:proofDJVP}.
In short, the inner $Lop$ represents the mapping of $\bv$ onto the linear operator $\J_{f}(\bx)^T$. Hence, ${\rm Lop}(f,\bx,\bv)^T$ being linear in $\bv$, can itself be differentiated w.r.t. $\bv$ and evaluated with direction $\bu$ leading to the above equality. While this alternative strategy might not be of great interest for theoretical studies since the asymptotic time-complexity of ${\rm Lop}$ and ${\rm Rop}$ is similar (proportional to the number of operations $K$ \citep{griewank2008evaluating}). Which of the above strategies will provide the fastest computation in practice depends on the type of operations (input/output shapes) that are present in the computational graph. And while Rop and Lop can be combined, the general problem of optimal Jacobian accumulation (finding the most efficient computation) is NP-complete \citep{naumann2008optimal}. Note that current software/hardware in deep learning heavily focus on Lop as backpropagation (gradient based DN training strategy) is a special case of the latter \citep{rumelhart1986learning}.

{\bf Deep Networks (DNs).}~A DN is an operator $f$ that maps an {\em observation} \footnote{We consider vectors, matrices and tensors as flattened vectors} $\bz \in \R^D$ to a {\em prediction} $\by \in \R^K$. This prediction is a nonlinear transformation of the input $\bx$ built through a cascade of linear and nonlinear operators \citep{lecun2015deep}.
Those operators include affine operators such as the {\em fully connected operator} (simply an affine transformation defined by weight matrix $\bW^{\ell}$ and bias vector $\bv^{\ell}$), {\em convolution operator} (with circulant $\bW^{\ell}$), and, nonlinear operators such as the {\em activation operator} (applying a scalar nonlinearity such as the ubiquitous ReLU), or the {\em max-pooling operator}; precise definitions of these operators can be found in \citet{goodfellow2016deep}.

It will become convenient for our development to group the succession of operators forming a DN into {\em layers} in order to express the entire DN as a composition of $L$ intermediate layers $f^{\ell}$, $\ell=1,\dots,L$ as in
\begin{align}
    f(\bx)=(f^{L}\circ \dots \circ f^{1})(\bx),
\end{align}
where each $f^{\ell}$ maps an input vector of length $D^{\ell-1}$ to an output vector of length $D^{\ell}$. We formally specify how to perform this grouping to ensure that any DN as a unique layer decomposition.

\begin{defn}[Deep Network Layer]
\label{def:layer}
A DN layer $f^{\ell}$ composes a single nonlinear operator and any (if any) of the preceding linear operators between it and the preceding nonlinear operator (if any).
\end{defn}
\qq

We also explicitly denote the up-to-layer $\ell$ mapping as
\begin{align*}
    \bz^{\ell}(\bx)=f^{\ell}\left(\bz^{\ell-1}(\bx)\right),
\end{align*}
with initialization $\bz^{0}(\bx)\triangleq \bx$, those intermediate representations $\bz^{1},\dots,\bz^{L-1}$ are denoted as {\em feature maps}.

{\bf Continuous Piecewise Affine (CPA) DNs.}~Whenever a DN $f$ employs CPA nonlinearities throughout its layers, the entire input-output mapping is a (continuous) affine spline with a partition $\Omega$ of its domain (the input space of the DN) and a per-region affine mapping (for more background on splines we refer the reader to \citet{de1978practical,schumaker2007spline,bojanov2013spline,pascanu2013number}) such that 
\begin{align}
    f(\bx)=&\sum_{\omega\in\Omega}\left(\bA_{\omega}\bx+\bb_{\omega}\right)\Indic_{\bx \in \omega}, \;\;\;\;\;\bx \in \mathbb{R}^{D},
    \label{eq:DNmapping}
\end{align}
where the per-region slope and bias parameters are a function of the DN architecture and parameters. Given an observation $\bx \in \omega$, the affine parameters of region $\omega$ from (\ref{eq:DNmapping}) can be obtained explicitly as follows.
\\
{\bf 1.}~For each layer $f^{\ell}$ (recall Def~\ref{def:layer}) compose the possibly multiple linear operator into a single one with slope matrix $\bW^{\ell}$ and bias $\bb^{\ell}$ (any composition of linear operators can be expressed this way by composition and distributive law) and denote the layer's {\em pre-activation} as
\begin{align}
    \bh^{\ell}(\bx)=\bW^{\ell}\bz^{\ell-1}(\bx)+\bb^{\ell},\label{eq:h}
\end{align}
in the absence of linear operators in a layer, set $\bW^{\ell}$ as the identity matrix and $\bb^{\ell}$ as the zero-vector.\\
{\bf 2.}~For each layer, encode the {\em states} of the layer's activation operator into the following diagonal matrix
\begin{align}
    \left[\bQ^{\ell}\left(\bh^{\ell}(\bx)\right)\right]_{i,i}=
    \begin{cases}
    \eta & \text{ if } \left[\bh^{\ell}(\bx)\right]_i < 0\\
    1 & \text{ else }
    \end{cases},\label{eq:Q}
\end{align}
where $\eta >0$ (leaky-ReLU), $\eta = 0$ (ReLU), or $\eta=-1$ (absolute value); for the max-pooling operator, see Appendix~\ref{appendix:maxpooling}; $[.]$ is used to access a specific entry of a matrix/vector, for clarity we will often omit the double superscript in (\ref{eq:Q}) and use $\bQ\left(\bh^{\ell}(\bx)\right)$.\\
{\bf 3.}~Combining (\ref{eq:h}) and (\ref{eq:Q}) to obtain
\begin{align*}
    \bA_{\omega} =& \prod_{\ell=0}^{L-1}\bQ\left(\bh^{L-\ell}(\bx)\right)\bW^{L-\ell},\\
    \bb_{\omega}=& \sum_{i=1}^{L}\left(\prod_{\ell=0}^{L-i-1}\bQ\left(\bh^{L-\ell}(\bx)\right)\bW^{L-\ell}\right)\bQ\left(\bh^{i}(\bx)\right)\bb^{i},
\end{align*}
with $\bA_{\omega} \in \mathbb{R}^{K\times D}$ and $\bb_{\omega} \in \mathbb{R}^{D}$.\\
Recently, many studies have successfully exploited the CPA property of current DNs to derive novel insights and results such as bounding the number of partition regions \citep{montufar2014number,arora2016understanding,rister2017piecewise,serra2018bounding,hanin2019complexity}, characterizing the geometry of the partitioning \citep{balestriero2019geometry,robinson2019dissecting,gamba2020hyperplane}; and quantifying the expressive power of CPA DNs as function approximators \citep{unser2019representer,bartlett2019nearly,nadav2020expression,grigsby2020transversality}. 
We propose to follow the same path and exploit the CPA property of DNs to obtain a fast and architecture independent method to evaluate Jacobian-vector products with DNs that do not require any flavor of automatic differentiation. Our method will allow to speed-up recent methods employing JVPs and to ease their deployment.

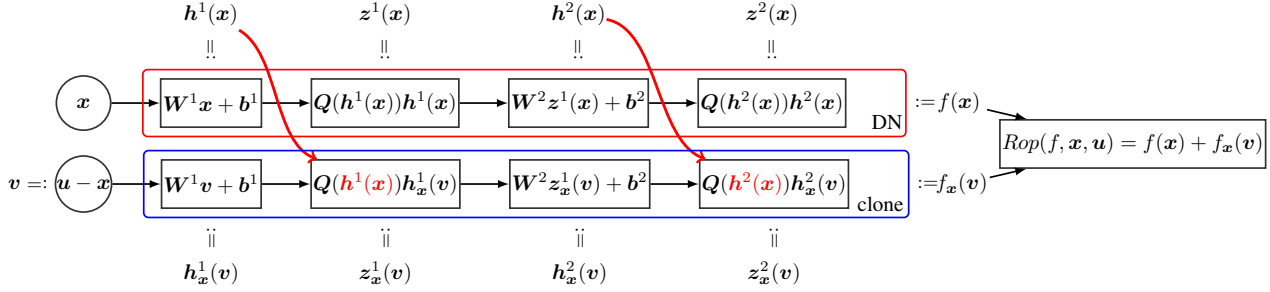
\begin{figure*}[t!]
    \centering
\resizebox{1\textwidth}{!}{
\begin{tikzpicture}[object/.style={thin,double,<->}]
\tikzstyle{main}=[rectangle, minimum width = 16.3mm, minimum height = 8mm, thick, draw =black!80, node distance = 5mm, inner xsep=1pt, inner ysep=0pt]
\tikzstyle{main2}=[circle, minimum size = 9mm, thick, draw =black!80, node distance = 5mm,inner sep=0pt]
\tikzstyle{main3}=[rectangle, minimum size = 9mm, thick,  node distance = 5mm,inner sep=0pt]
\tikzstyle{connect}=[-latex, thick]
\tikzstyle{box}=[rectangle, draw=black!100]

\node[main2](x) []{$\bx$};
\node[main2](xtilde) [below=0.4cm of x]{$\bu-\bx$};

\node[main] (h1) [right=0.8cm of x] {$\bW^{1}\bx+\bb^{1}$};
\node[main] (h1tilde) [right=0.8cm of xtilde] {$\bW^{1}\bv+\bb^{1}$};

\node[main] (z1) [right=0.8cm of h1] {$\bQ(\bh^{1}(\bx))\bh^{1}(\bx)$};
\node[main] (z1tilde) [right=0.8cm of h1tilde] {$\bQ($\textcolor{red}{$\bh^{1}(\bx)$}$)\bh^{1}_{\bx}(\bv)$};

\node[main] (h2) [right=0.8cm of z1] {$\bW^{2}\bz^{1}(\bx)+\bb^{2}$};
\node[main] (h2tilde) [right=0.8cm of z1tilde] {$\bW^{2}\bz^{1}_{\bx}(\bv)+\bb^{2}$};

\node[main] (z2) [right=0.8cm of h2] {$\bQ(\bh^{2}(\bx))\bh^{2}(\bx)$};
\node[main] (z2tilde) [right=0.8cm of h2tilde] {$\bQ($\textcolor{red}{$\bh^{2}(\bx)$}$)\bh^{2}_{\bx}(\bv)$};

\node[main3](f) [right=1.4cm of z2]{$f(\bx)$};
\node[main3](ftilde) [right=1.4cm of z2tilde]{$f_{\bx}(\bv)$};

\node[main] (z3) [below right=-0.2cm and 0.2cm of f] {$Rop(f,\bx,\bu)=f(\bx)+f_{\bx}(\bv)$};

\node[main3] (H1) [above=0.6cm of h1] {$\bh^{1}(\bx)$};
\node[main3] (Z1) [above=0.6cm of z1] {$\bz^{1}(\bx)$};
\node[main3] (H2) [above=0.6cm of h2] {$\bh^{2}(\bx)$};
\node[main3] (Z2) [above=0.6cm of z2] {$\bz^{2}(\bx)$};
\node[main3] (V) [left=0.2cm of xtilde] {$\bv$};

\node[main3] (H1tilde) [below=0.6cm of h1tilde] {$\bh^{1}_{\bx}(\bv)$};
\node[main3] (Z1tilde) [below=0.6cm of z1tilde] {$\bz^{1}_{\bx}(\bv)$};
\node[main3] (H2tilde) [below=0.6cm of h2tilde] {$\bh^{2}_{\bx}(\bv)$};
\node[main3] (Z2tilde) [below=0.6cm of z2tilde] {$\bz^{2}_{\bx}(\bv)$};

\node (1) at ($(h1)!0.6!(H1)$) {\rotatebox{90}{{\small $:=$}}};
\node (2) at ($(z1)!0.6!(Z1)$) {\rotatebox{90}{{\small $:=$}}};
\node (3) at ($(z2)!0.6!(Z2)$) {\rotatebox{90}{{\small $:=$}}};
\node (4) at ($(h2)!0.6!(H2)$) {\rotatebox{90}{{\small $:=$}}};
\node (1) at ($(h1tilde)!0.6!(H1tilde)$) {\rotatebox{270}{{\small $:=$}}};
\node (2) at ($(z1tilde)!0.6!(Z1tilde)$) {\rotatebox{270}{{\small $:=$}}};
\node (3) at ($(z2tilde)!0.6!(Z2tilde)$) {\rotatebox{270}{{\small $:=$}}};
\node (4) at ($(h2tilde)!0.6!(H2tilde)$) {\rotatebox{270}{{\small $:=$}}};
\node (4) at ($(V)!0.33!(xtilde)$) {\rotatebox{180}{{\small $:=$}}};

\node (2) at ($(z2)!0.82!(f)$) {\rotatebox{0}{{\small $:=$}}};
\node (3) at ($(z2tilde)!0.82!(ftilde)$) {\rotatebox{0}{{\small $:=$}}};

\path (x) edge [connect] (h1)
      (xtilde) edge [connect](h1tilde)
      (h1) edge [connect] (z1)
      (h1tilde) edge [connect](z1tilde)
      (z1) edge [connect] (h2)
      (z1tilde) edge [connect](h2tilde)
      (h2) edge [connect] (z2)
      (h2tilde) edge [connect](z2tilde)
      (f) edge [connect] (z3)
      (ftilde) edge [connect](z3)
		;
\draw[->,line width=1.34pt,red, looseness=0.8]
  (H1) to[out=-30,in=160] (z1tilde);
\draw[->,line width=1.34pt,red,looseness=0.8]
  (H2) to[out=-11,in=160] (z2tilde);

  \node[plate3=DN, xshift=10pt,inner sep=4pt,inner xsep=18pt, fit=(h1)(z2),draw=red] (plate1) {};
  \node[plate=clone, xshift=10pt, inner sep=4pt,inner xsep=18pt, fit=(h1tilde)(z2tilde),draw=blue] (plate1) {};
\end{tikzpicture}
}
\vspace{-0.8cm}
    \caption{ \small
    Depiction of the DN cloning strategy employed in (\ref{eq:JVPclone}) to evaluate Jacobian-vector of a DN taken w.r.t the DN input, at $\bx$, and with direction $\bu$ without resorting to automatic differentiation. First, in red, one propagates the input $\bx$ and observes the nonlinearities' states (recall (\ref{eq:Q})). Then, in blue, one feeds $\bv :=\bu-\bx$ in the exact same DN but fixing the nonlinearities' states to the ones observed during the forward pass of $\bx$ (blue). The feature map of layer $\ell$ obtained when fixing the nonlinearities from $\bx$ and forwarding $\bv$ is $\bz^{\ell}_{\bx}(\bv)$ (and similarly $\bh^{\ell}_{\bx}(\bv)$ for the pre-activations). The produced outputs, $f_{\bx}(\bv)$ and $f(\bx)$, can then be combined as per (\ref{eq:JVPclone}) to obtain the desired Jacobian-vector product.
    }
    \label{fig:cloning}
\end{figure*}


\section{Fast Jacobian-Vector Product for Continuous Piecewise Affine Deep Networks}
\label{sec:fastJVP}

As we mentioned, the core of our solution resides on the assumptions that the employed DN's nonlinearities are CPA, we first describe our solution, its implementation and then conclude with thorough empirical validations and solutions to extend our technique to smooth DNs.

\subsection{Deep Network Cloning}

When employing CPA nonlinearities, the DN input-output mapping can be expressed as a per-region affine mapping (recall (\ref{eq:DNmapping})). This key property greatly simplifies the form of the  explicit Rop (or JVP) of a DN taken with respect to its input given an observation $\bx$ and a directional vector $\bu$. In fact, we directly have
\begin{align}
    {\rm Rop}(f,\bx,\bu)=\bA_{\omega(\bx)}\bu, \;\;\omega(\bx) \text{ s.t. } \bx \in \omega,\omega\in\Omega.\label{eq:explicit}
\end{align}
where we explicit the $\bx$ enclosing region $\omega$ with $\omega(\bx)$.
In the unrealistic setting where one would a priori know and store the $\bA_{\omega}$ matrices for each region $\omega \in \Omega$, and instantaneously infer the region $\omega(\bx)$, the evaluation of (\ref{eq:explicit}) would be performed in $\mathcal{O}(KD)$ time complexity. This complexity could be reduced in the presence of low-rank (or other peculiar cases) $\bA_{\omega}$ matrices,  such cases do not occur with standard architectures and are thus omitted. In the more realistic scenario however, direct evaluation of \ref{eq:explicit} is not possible as each matrix tends to be large (and not sparse) with input dimension $D \gg 3000$, and outpud dimension $K$ reaching $1000$ for classification/regression tasks, and being as large as $D$ for the case of autoencoders; and the number of regions $Card(\Omega)$ being on the order of hundreds of thousands \cite{cosentino2020provable}, even for small DNs (and keep increasing with the DN number of nodes/layers). As a result, all current solutions rely on automatic differentiation which constructs only a single $\bA_{\omega(\bx)}$ and $\bb_{\omega(\bx)}$ on demand based on the given observation $\bx$ and its propagation through the DN layers.

Instead, we propose an alternative solution that does not rely on automatic differentiation. For clarity, we detail our approach first when the JVPs is taken with respect to the DN input, and second when the JVP is taken with respect to the parameters of a layer.

{\bf JVPs of DN inputs.}~Recalling (\ref{eq:DNmapping}), we know that forwarding an observation $\bx$ through the DN layers produces the output $\bA_{\omega}\bx+\bb_{\omega}$.
Now, suppose that we possessed the ability to {\em freeze}, or {\em clone} the layers such that the internal $\bQ^{\ell}(\bh^{\ell}(\bx))$ matrices (recall \ref{eq:Q}) remain the same regardless of the presented input; denote such a DN cloned at $\bx$ by $f_{\bx}$. Then, in all simplicity by using (\ref{eq:DNmapping}) we obtain the following equality
\begin{align}
    f(\bx)-f_{\bx}(\bu-\bx)=&\bA_{\omega(\bx)}\bx+\bb_{\omega(\bx)}\nonumber\\
    &-\left(\bA_{\omega(\bx)}(\bu-\bx)+\bb_{\omega(\bx)}\right)\nonumber\\
    =&\bA_{\omega(\bx)}\bu = {\rm Rop}(f,\bx,\bu),\label{eq:JVPclone}
\end{align}
producing the desired JVP. {\em Note that the user only needs to perform the forward pass through the cloned DN $f_{\bx}$, without the need to explicit compute $\bA_{\omega(\bx)}$, the JVP $\bA_{\omega(\bx)}\bu$ is gradually computed through the forwaard pass}.
Computationally, (\ref{eq:JVPclone}) requires only two passes through the DN that we group into two steps. Step 1: forward the observation $\bx$ through all the layers to observe the $\bQ^{\ell}(\bh^{\ell}(\bx))$ matrices. Step 2: evaluate the cloned DN at $\bu-\bx$ using the observed $\bQ^{\ell}(\bh^{\ell}(\bx))$ matrices from step 1.
We depict this cloning and propagation procedure in Fig.~\ref{fig:cloning}.

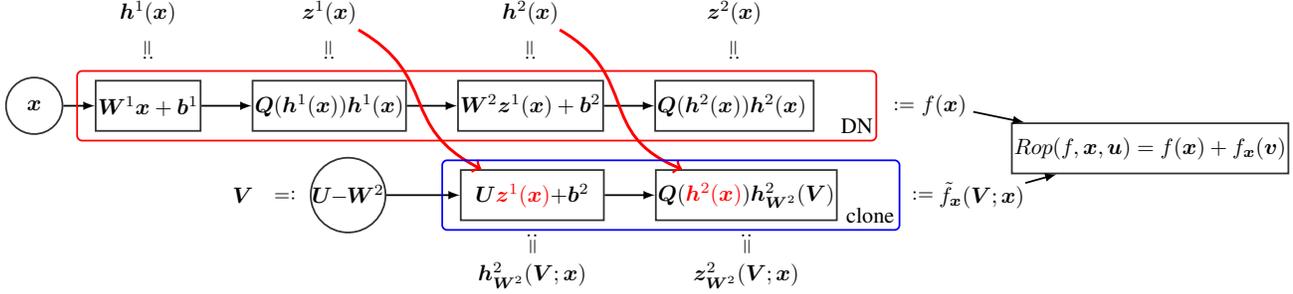
\begin{figure*}[t!]
    \centering
\resizebox{1\textwidth}{!}{
\begin{tikzpicture}[object/.style={thin,double,<->}]
\tikzstyle{main}=[rectangle, minimum width = 16.3mm, minimum height = 8mm, thick, draw =black!80, node distance = 5mm, inner xsep=1pt]
\tikzstyle{main2}=[circle, minimum size = 9mm, thick, draw =black!80, node distance = 5mm,inner sep=0pt]
\tikzstyle{main3}=[rectangle, minimum size = 9mm, thick, node distance = 5mm,inner sep=0pt]
\tikzstyle{connect}=[-latex, thick]
\tikzstyle{box}=[rectangle, draw=black!100]

\node[main2](x) []{$\bx$};
\node[main2](xtilde) [below=0.4cm of z1]{$\bU\hspace{-0.1cm}-\hspace{-0.1cm}\bW^{2}$};
\node[main3](V) [left=0.6cm of xtilde]{$\bV$};

\node[main] (h1) [right=0.5cm of x] {$\bW^{1}\bx+\bb^{1}$};

\node[main] (z1) [right=0.8cm of h1] {$\bQ(\bh^{1}(\bx))\bh^{1}(\bx)$};

\node[main] (h2) [right=0.8cm of z1] {$\bW^{2}\bz^{1}(\bx)+\bb^{2}$};
\node[main] (h2tilde) [right=1.17cm of xtilde] {$\hspace{0.2cm}\bU$\textcolor{red}{$\bz^{1}(\bx)$}$+\bb^{2}\hspace{0.2cm}$};

\node[main] (z2) [right=0.8cm of h2] {$\bQ(\bh^{2}(\bx))\bh^{2}(\bx)\hspace{0.08cm}$};
\node[main] (z2tilde) [right=0.8cm of h2tilde] {$\bQ($\textcolor{red}{$\bh^{2}(\bx)$}$)\bh^{2}_{\bW^{2}}(\bV)$};

\node[main3](f) [right=1.6cm of z2]{$f(\bx)$};
\node[main3](ftilde) [right=1.6cm of z2tilde]{$\tilde{f}_{\bx}(\bV;\bx)$};

\node[main3] (H1) [above=0.6cm of h1] {$\bh^{1}(\bx)$};
\node[main3] (Z1) [above=0.6cm of z1] {$\bz^{1}(\bx)$};
\node[main3] (H2) [above=0.6cm of h2] {$\bh^{2}(\bx)$};
\node[main3] (Z2) [above=0.6cm of z2] {$\bz^{2}(\bx)$};
\node[main3] (H2tilde) [below=0.4cm of h2tilde] {$\bh^{2}_{\bW^{2}}(\bV;\bx)$};
\node[main3] (Z2tilde) [below=0.4cm of z2tilde] {$\bz^{2}_{\bW^{2}}(\bV;\bx)$};

\node (1) at ($(h1)!0.6!(H1)$) {\rotatebox{90}{{\small $:=$}}};
\node (2) at ($(z1)!0.6!(Z1)$) {\rotatebox{90}{{\small $:=$}}};
\node (3) at ($(z2)!0.6!(Z2)$) {\rotatebox{90}{{\small $:=$}}};
\node (4) at ($(h2)!0.6!(H2)$) {\rotatebox{90}{{\small $:=$}}};
\node (1) at ($(h2tilde)!0.6!(H2tilde)$) {\rotatebox{270}{{\small $:=$}}};
\node (2) at ($(z2tilde)!0.6!(Z2tilde)$) {\rotatebox{270}{{\small $:=$}}};
\node (4) at ($(V)!0.4!(xtilde)$) {\rotatebox{180}{{\small $:=$}}};

\node (2) at ($(z2)!0.81!(f)$) {\rotatebox{0}{{\small $:=$}}};
\node (3) at ($(z2tilde)!0.75!(ftilde)$) {\rotatebox{0}{{\small $:=$}}};

\node[main] (z3) [below right=-0.2cm and 0.6cm of f] {$Rop(f,\bx,\bu)=f(\bx)+f_{\bx}(\bv)$};

\path (x) edge [connect] (h1)
      (xtilde) edge [connect](h2tilde)
      (h1) edge [connect] (z1)
      (z1) edge [connect] (h2)
      (h2) edge [connect] (z2)
      (h2tilde) edge [connect](z2tilde)
      (f) edge [connect] (z3)
      (ftilde) edge [connect] (z3)
		;
\draw[->,line width=1.34pt,red, looseness=1.3]
  (H2) to[out=-30,in=158] (z2tilde);
\draw[->,line width=1.34pt,red, looseness=1.3]
  (Z1) to[out=-30,in=153] (h2tilde);

  \node[plate3=DN, xshift=10pt,inner sep=4pt,inner xsep=18pt, fit=(h1)(z2),draw=red] (plate1) {};
  \node[plate=clone, xshift=10pt, inner sep=4pt,inner xsep=18pt, fit=(h2tilde)(z2tilde),draw=blue] (plate1) {};
\end{tikzpicture}
}
\vspace{-0.8cm}
    \caption{ \small
    Reprise of Fig.~\ref{fig:cloning} but now evaluating the Jacobian-vector of a DN at $\bx$ taken w.r.t the layer parameter $\bW^{2}$ and with direction $\bU$. First, one propagates the input $\bx$ and observes the nonlinearities' states (recall (\ref{eq:Q})). Then, one (starting from $\bz^{\ell-1}(\bx)$ propagates $\bx$ in the cloned DN by using $\bU-\bW^{\ell}$ instead of $\bW^{\ell}$. The produced outputs can then be combined as per (\ref{eq:JVPclone2}) to obtain the desired Jacobian-vector product ${\rm Rop}(f(\bx),\bW^{\ell}, \bU)$.
    }
    \label{fig:cloning2}
\end{figure*}

{\bf JVP of DN layer weights.}~We now consider the evaluation of a DN's JVP taken with respect to one of the layers' affine parameter i.e., weights, say $\bW^{\ell}$ for some layer $1\leq \ell \leq L$. Let's first define an auxiliary DN mapping $\tilde{f}(.;\bx)$ that takes as input the value of the weight of interest ($\bW^{\ell}$ in this case). This leads to the following mapping
\begin{align}
    \tilde{f}(\bW;\bx)=&(f^{L}\circ \dots \circ f^{\ell+1}\circ \sigma)(\bW\bz^{\ell}(\bx)+\bb^{\ell})\nonumber\\
    =(f^{L}&\circ \dots \circ f^{\ell+1}\circ \sigma)(\bM(\bz^{\ell}(\bx))\bw+\bb^{\ell})\label{eq:equalM}
\end{align}
where $\bw$ is the flattened version of $\bW$ and $\bM(\bz^{\ell}(\bx))$ is the matrix s.t. projecting $\bw$ onto it is equivalent to projecting $\bz^{\ell}(\bx)$ onto $\bW$. If $\bW$ is a dense matrix then $\bM$ is block diagonal with $D^{\ell}$ rows and $D^{\ell}D^{\ell-1}$ columns as
$$\bM(\bz^{\ell}(\bx))=\begin{pmatrix}
\bz^{\ell}(\bx)^T&0&\dots\\
0&\bz^{\ell}(\bx)^T&\dots\\
\vdots&\vdots&\ddots
\end{pmatrix},$$
if $\bW$ were a convolutional layer parameter, then the $\bM$ matrix would include the patch-translations corresponding to the application of the filter. Now, it is clear that taking the JVP of $\tilde{f}(\bW;\bx)$ w.r.t. $\bW$ can be done exactly as per (\ref{eq:JVPclone}) since the flattened layer parameter can be seen as the DN input leading to
\begin{align}
    \tilde{f}(\bW^{\ell};\bx)-\tilde{f}_{\bW^{\ell}}(\bU-\bW^{\ell};\bx)= {\rm Rop}(f(\bx),\bW^{\ell},\bU),\label{eq:JVPclone2}
\end{align}
where $\bU$ is the the direction of the JVP that has same shape as $\bW^{(\ell)}$. Also, in this case, $\tilde{f}_{\bW^{\ell}}$ represents cloning the nonlinearity states $\bQ^{(\ell)}$ from the DN with inputs $\bW^{(\ell)}$ and $\bx$. Furthermore, thanks to the equality of (\ref{eq:equalM}), one does not even need to actually obtain $\bM$ since performing the usual layer forward propagation with the given parameter and the layer input is equivalent. As a result, regardless of the type of layer, taking the JVP w.r.t. a layer parameter is done simply by (i) propagating $\bx$ through the DN $f$ while using $\bW^{(\ell)}$ to observe all the matrices $\bQ^{\ell}(\bh^{\ell}(\bx))$, (ii) propagating $\bx$ using $\bU-\bW^{\ell}$ as the filter of the layer's parameter of interest (and keeping all the nonlinearities frozen) and (iii) computing the different between the two produced quantity.  We depict this cloning procedure in Fig.~\ref{fig:cloning2}.

We now formalize those two cases in the following theorem that supports the main results of our study, the proof as well as many additional examples and details is provided in Appendix \ref{proof:JVP}, implementation details and experiments are provided in the following sections.

\begin{thm}
\label{thm:JVP}
Evaluating the JVP of a DN w.r.t. its input at $\bx$ with direction $\bu$ (recall (\ref{eq:rop})) and evaluating the JVP of a DN at $\bW^{\ell},\bx$ w.r.t. its layer parameter with direction $\bU$ can be done with 
\begin{align*}
{\rm Rop}(f,\bx,\bu)=&f(\bx)-f_{\bx}(\bu-\bx),\\
{\rm Rop}(f(\bx),\bW,\bU)=&\tilde{f}(\bW^{\ell};\bx)-\tilde{f}_{\bW^{\ell}}(\bU-\bW^{\ell};\bx).
\end{align*}
\end{thm}

Note that in practice, the two forward passes are done synchronously in a single pass by concatenating the two inputs in a single mini-batch. Leveraging the highly parallel computations enabled by GPUs, Thm.~\ref{thm:JVP} enables rapid evaluation of JVPs without requiring any automatic differentiation to be performed as we will detail in the following section.

\subsection{Implementation}
\label{sec:implementation}

\begin{algorithm}[t!]
\caption{{\small Custom nonlinearity implementations to be used jointly with the DN input that concatenates $\bx,\bu,\mathbf{0}$ (recall (\ref{eq:JVPclone}) and (\ref{eq:JVPclone2})). Those implementations are drop-in replacements of the usual \pyth{tf.nn.relu} (leakiness to $0$), \pyth{tf.nn.leaky_relu} (leakiness $>0$), \pyth{tf.abs} (leakiness to $-1$). One can use our method to evaluate JVPs only by using those custom nonlinearities in place of the usual ones and feed to the DN the concatenated inputs $X$, no additional changes are required. Codes for additional nonlinearities e.g. max-pooling, dropout, as well as smooth nonlinearity such as batch-normalization, which is a smooth nonlinear operator during training, are provided in Appendix~\ref{sec:codes}}}
\label{algo:implementation}
\begin{python}
def clone_act(X, leak):
  x_only = X[: X.shape[0]//2]
  mask = tf.greater(x_only, 0)
  tmask = tf.tile(mask, [2]+[1]*(X.ndim-1))
  return tf.where(tmask, 1.0, leak) * X
\end{python}
\end{algorithm}

We describe in detail the implementation of our method which only takes the form of a few lines of code, we focus here on taking the JVP of a DN with respect to its input given by (\ref{eq:JVPclone}) and using Tensorflow \cite{tensorflow2015-whitepaper} for concreteness and refer the reader to Appendix~\ref{sec:layer_implementation} for the case of JVP taken with respect to layer weights (\ref{eq:JVPclone2}) and for further details and discussions on the implementation.

As was explained in the previous section, our method requires to forward-pass three different DN inputs $\bx,\bu,\mathbf{0}$ (recall (\ref{eq:JVPclone})) and to clone the nonlinearities ($\bQ^{\ell}(\bh^{\ell}(\bx))$) for the forward-pass of $\bu,\mathbf{0}$ from the forward-pass of $\bx$. This is implemented as follows
\begin{enumerate}[noitemsep,nolistsep,topsep=0pt,leftmargin=10pt]
    \item Given a DN of interest $f$, replace all the nonlinearities (e.g. \pyth{tf.nn.relu} and \pyth{tf.nn.max_pool}) with the custom ones provided in Algo.~\ref{algo:implementation} leading to a custom DN denoted as $F$
    \item Propagate through this custom DN $F$ a single input $X$ formed as \pyth{X=tf.concat([x,u,tf.zeros_like(x)], 0)} that concatenates the three needed inputs over the mini-batch dimension, the shape of $\bx,\bu$ and $\mathbf{0}$ must be identical, one can process a single or multiple JVPs at once based on the shape of those inputs
    \item Obtain the needed quantities to compute the JVP from (\ref{eq:JVPclone}) with $f_{\bx}(\bu)$\pyth{=F(X)[len(x):2*len(x)]} and $f_{\bx}(\mathbf{0})$\pyth{=F(X)[2*len(x):]}, the usual DN output is computed as a side effect $f(\bx)$\pyth{=F(X)[:len(x)]}.
\end{enumerate}

\begin{table*}[t!]
    \centering
    \setlength\tabcolsep{3pt} 
    \renewcommand{\arraystretch}{0.8}
    \caption{\small Average computation time to evaluate the JVP ${\rm Rop}(f,\bx,\bu)$, averaged over $1000$ runs for $13$ different DN architectures, $2$ different input/output shapes and $2$ different GPUs. For the RNN case, the recurrence is taken along the horizontal axis of the image. The symbol `-' indicates Out Of Memory (OOM) failure. We compare our method (clone) against `batch jacobian' (\ref{eq:JVPtimes}), `jvp` which is the forward-mode auto differentiation, and 'double vjp' (\ref{eq:doubleVJP}). See Appendix~\ref{sec:details} for details on implementations, libraries and hardware. We observe that our method produces faster JVP evaluation in all cases with speed-up varying based on the architecture at hand which are consistent across input/output shapes and hardware. For GeForce GTX 1080Ti times, see Table~\ref{tab:timesXcpu} where the same trend can be observed, for standard deviations, see Table~\ref{tab:std_x}, in the Appendix.}
    \begin{tabular}{c|l|l||rrrr|r||rrrr|r|}
        \cline{4-13}
        \multicolumn{1}{r}{}&\multicolumn{2}{r|}{}&\multicolumn{5}{c||}{$\bx \in \mathbb{R}^{100\times 100 \times 3},\by\in\mathbb{R}^{20}$}&\multicolumn{5}{c|}{$\bx \in \mathbb{R}^{400\times 400 \times 3},\by\in\mathbb{R}^{1000}$}\\ \cline{2-13}
        &Model & \# params. & \makecell{batch-\\jacobian} & jvp & \makecell{double \\ vjp} & \makecell{clone \\ (ours)} & \makecell{ speedup \\ factor} & \makecell{batch-\\jacobian} & jvp & \makecell{double \\ vjp} & \makecell{clone \\ (ours)}&\makecell{ speedup \\ factor}\\ \toprule
        \multirow{13}{*}{\rotatebox{90}{Quadro RTX 8000}}&RNN &5M& 0.264 & - & 0.233 & 0.036 & 6.47 & - & - & 0.900 & 0.129 & 6.98 \\
        \cline{2-13}
        &UNet &49M& - & 0.047 & 0.390 & 0.024 & 1.63 & - & 0.059 & 0.037 & 0.022 & 1.68 \\
        \cline{2-13}
         &VGG16 &138M& 0.038 & 0.014 & 0.014 & 0.005 & 2.80 & - & 0.168 & 0.158 & 0.046 & 3.43 \\
        \cline{2-13}
        &VGG19&143M&0.048 & 0.020 & 0.017 & 0.006 & 2.83 & - & 0.210 & 0.193 & 0.059 & 3.27 \\ \cline{2-13}
        &Inception V3&23M&0.037 & 0.030 & 0.025 & 0.016 & 1.56 & - & 0.028 & 0.027 & 0.019 & 1.42 \\\cline{2-13}
        &Resnet50&25M&0.030 & 0.019 & 0.015 & 0.012 & 1.25 & - & 0.028 & 0.020 & 0.018 & 1.12 \\\cline{2-13}
        &Resnet101&44M&0.051 & 0.029 & 0.024 & 0.022 & 1.09 & - & 0.038 & 0.036 & 0.030 & 1.13 \\ \cline{2-13}
        &Resnet152&60M&0.680 & 0.037 & 0.039 & 0.029 & 1.28 & - & 0.053 & 0.049 & 0.036 & 1.36 \\ \cline{2-13}
        &EfficientNet B0 &5M&0.053 & 0.023 & 0.018 & 0.013 & 1.38 & - & 0.032 & 0.021 & 0.015 & 1.40 \\ \cline{2-13}
        &EfficientNet B1 &7M&0.073 & 0.028 & 0.024 & 0.018 & 1.33 & - & 0.032 & 0.025 & 0.019 & 1.32 \\ \cline{2-13}
        &Densenet121&8M&0.040 & 0.036 & 0.033 & 0.022 & 1.50 & - & 0.036 & 0.033 & 0.025 & 1.32 \\ \cline{2-13}
        &Densenet169&14M&0.054 & 0.046 & 0.034 & 0.026 & 1.31 & - & 0.047 & 0.037 & 0.028 & 1.32\\\cline{2-13}
        &Densenet202&17M&0.070 & 0.056 & 0.042 & 0.033 & 1.27 & - & 0.055 & 0.043 & 0.033 & 1.30\\ \bottomrule
        \multicolumn{1}{r}{}&\multicolumn{6}{r}{arithmetic average speedup from fastest alternative:}&\multicolumn{1}{c}{{\bf 1.98}}&\multicolumn{4}{r}{}&\multicolumn{1}{c}{{\bf 2.08}} \\
        \multicolumn{1}{r}{}&\multicolumn{6}{r}{geometric average speedup from fastest alternative:}&\multicolumn{1}{c}{{\bf 1.71}}&\multicolumn{4}{r}{}&\multicolumn{1}{c}{{\bf 1.74}} \\\bottomrule
        \multirow{13}{*}{\rotatebox{90}{TITAN X (Pascal)}}&RNN &5M& - & - & 0.220 & 0.039 & 5.64 & - & - & - & 0.134 & $\infty$ \\ \cline{2-13}
        &UNet &49M& - & 0.083 & 0.069 & 0.04 & 1.72 & - & 0.118 & 0.084 & 0.057 & 1.47 \\
        \cline{2-13}
         &VGG16 &138M& 0.059 & 0.026 & 0.019 & 0.012 & 1.58 & - & 0.122 & 0.088 & 0.062 & 1.42 \\
        \cline{2-13}
        &VGG19&143M&0.072 & 0.034 & 0.026 & 0.017 & 1.53 & - & 0.133 & 0.101 & 0.077 & 1.31 \\ \cline{2-13}
        &Inception V3&23M&0.043 & 0.033 & 0.028 & 0.017 & 1.65 & - & 0.038 & 0.033 & 0.019 & 1.74 \\\cline{2-13}
        &Resnet50&25M&0.050 & 0.019 & 0.016 & 0.011 & 1.45 & - & 0.022 & 0.017 & 0.013 & 1.31 \\\cline{2-13}
        &Resnet101&44M&0.069 & 0.032 & 0.029 & 0.02 & 1.45 & - & 0.041 & 0.032 & 0.026 & 1.23 \\ \cline{2-13}
        &Resnet152&60M&0.091 & 0.051 & 0.041 & 0.028 & 1.46 & - & 0.057 & 0.046 & 0.033 & 1.39 \\ \cline{2-13}
        &EfficientNet B0 &5M&0.064 & 0.025 & 0.021 & 0.01 & 2.10 & - & 0.030 & 0.024 & 0.012 & 2.00 \\ \cline{2-13}
        &EfficientNet B1 &7M&0.086 & 0.035 & 0.028 & 0.014 & 2.00 & - & 0.040 & 0.033 & 0.016 & 2.06 \\ \cline{2-13}
        &Densenet121&8M&0.055 & 0.046 & 0.036 & 0.022 & 1.64 & - & 0.055 & 0.040 & 0.025 & 1.60 \\ \cline{2-13}
        &Densenet169&14M&0.077 & 0.071 & 0.049 & 0.029 & 1.69 & - & 0.076 & 0.055 & 0.033 & 1.67\\\cline{2-13}
        &Densenet202&17M&0.094 & 0.083 & 0.051 & 0.035 & 1.46 & - & 0.087 & 0.064 & 0.04 & 1.60\\ \bottomrule
        \multicolumn{1}{r}{}&\multicolumn{6}{r}{arithmetic average speedup from fastest alternative:}&\multicolumn{1}{c}{{\bf 1.95}}&\multicolumn{4}{r}{}&\multicolumn{1}{c}{{\bf 1.57}} \\
        \multicolumn{1}{r}{}&\multicolumn{6}{r}{geometric average speedup from fastest alternative:}&\multicolumn{1}{c}{{\bf 1.79}}&\multicolumn{4}{r}{}&\multicolumn{1}{c}{{\bf 1.49}} \\\bottomrule
    \end{tabular}
    \label{tab:timesX}
\end{table*}

\begin{table*}[t!]
    \centering
    \setlength\tabcolsep{3pt} 
    \renewcommand{\arraystretch}{0.8}
    \caption{\small Reprise of Table~\ref{tab:timesX} but now considering ${\rm Rop}(f(\bx),\bW,\bU)$. For GeForce GTX 1080Ti times, see Table~\ref{tab:timesWcpu} where the same trend can be observed, for standard deviations, see Table~\ref{tab:std_w}, in the Appendix.}
    \begin{tabular}{c|l|l||rrrr|r||rrrr|r|}
        \cline{4-13}
        \multicolumn{3}{r|}{}&\multicolumn{5}{c||}{$\bx \in \mathbb{R}^{100\times 100 \times 3},\by\in\mathbb{R}^{20}$}&\multicolumn{5}{c|}{$\bx \in \mathbb{R}^{400\times 400 \times 3},\by\in\mathbb{R}^{1000}$}\\ \cline{2-13}
        \multicolumn{1}{r|}{}&Model & \# params. & \makecell{batch-\\jacobian} & jvp & \makecell{double \\ vjp} & \makecell{clone \\ (ours)} & \makecell{ speedup \\ factor} & \makecell{batch-\\jacobian} & jvp & \makecell{double \\ vjp} & \makecell{clone \\ (ours)}&\makecell{ speedup \\ factor}\\ \toprule
        \multirow{13}{*}{\rotatebox{90}{Quadro RTX 8000}}&RNN&5M&0.245 & - & 0.227 & 0.047 & 4.83 & - & - & 0.877 & 0.163 & 5.38 \\ \cline{2-13}
        &UNet &49M&- & 0.048 & 0.04 & 0.023 & 1.74 & - & 0.062 & 0.042 & 0.022 & 1.91 \\ \cline{2-13}
         &VGG16 &138M& 0.036 & 0.014 & 0.014 & 0.005 & 2.80 & - & 0.168 & 0.157 & 0.047 & 3.34 \\
        \cline{2-13}
        &VGG19&143M&0.046 & 0.021 & 0.016 & 0.006 & 2.67 & - & 0.205 & 0.191 & 0.055 & 3.47 \\ \cline{2-13}
        &Inception V3&23M&0.032 & 0.033 & 0.023 & 0.017 & 1.35 & - & 0.031 & 0.026 & 0.018 & 1.44 \\\cline{2-13}
        &Resnet50&25M&0.027 & 0.019 & 0.015 & 0.012 & 1.25 & - & 0.028 & 0.021 & 0.019 & 1.11 \\\cline{2-13}
        &Resnet101&44M&0.051 & 0.034 & 0.026 & 0.021 & 1.24 & - & 0.038 & 0.032 & 0.029 & 1.10 \\ \cline{2-13}
        &Resnet152&60M& 0.059 & 0.039 & 0.035 & 0.028 & 1.25 & - & 0.052 & 0.042 & 0.035 & 1.20 \\ \cline{2-13}
        &EfficientNet B0 &5M&0.051 & 0.022 & 0.018 & 0.013 & 1.38 & - & 0.023 & 0.020 & 0.015 & 1.33 \\ \cline{2-13}
        &EfficientNet B1 &7M&0.072 & 0.028 & 0.024 & 0.016 & 1.50 & - & 0.028 & 0.027 & 0.018 & 1.50 \\ \cline{2-13}
        &Densenet121&8M&0.039 & 0.036 & 0.031 & 0.024 & 1.29 & - & 0.038 & 0.030 & 0.024 & 1.25 \\ \cline{2-13}
        &Densenet169&14M&0.062 & 0.055 & 0.035 & 0.028 & 1.25 & - & 0.048 & 0.035 & 0.028 & 1.25\\\cline{2-13}
        &Densenet201&17M&0.064 & 0.055 & 0.041 & 0.031 & 1.32 & - & 0.062 & 0.042 & 0.037 & 1.14\\ \bottomrule
        \multicolumn{1}{r}{}&\multicolumn{6}{r}{arithmetic average speedup from fastest alternative:}&\multicolumn{1}{c}{{\bf 1.84}}&\multicolumn{4}{r}{}&\multicolumn{1}{c}{{\bf 1.96}} \\ 
        \multicolumn{1}{r}{}&\multicolumn{6}{r}{geometric average speedup from fastest alternative:}&\multicolumn{1}{c}{{\bf 1.66}}&\multicolumn{4}{r}{}&\multicolumn{1}{c}{{\bf 1.69}} \\\bottomrule
        \multirow{13}{*}{\rotatebox{90}{TITAN X (Pascal)}}&RNN&5M&- & - & 0.219 & 0.045 & 4.87 & - & - & - & 0.153 & $\infty$ \\ \cline{2-13}
        &UNet &49M&- & 0.084 & 0.068 & 0.039 & 1.74 & - & 0.114 & 0.079 & 0.059 & 1.34 \\ \cline{2-13}
         &VGG16 &138M& 0.059 & 0.024 & 0.019 & 0.012 & 1.58 & - & 0.114 & 0.085 & 0.062 & 1.37 \\
        \cline{2-13}
        &VGG19&143M&0.072 & 0.032 & 0.025 & 0.016 & 1.56 & - & 0.137 & 0.103 & 0.089 & 1.16 \\ \cline{2-13}
        &Inception V3&23M&0.041 & 0.033 & 0.029 & 0.017 & 1.71 & - & 0.039 & 0.035 & 0.021 & 1.67 \\\cline{2-13}
        &Resnet50&25M&0.048 & 0.019 & 0.015 & 0.011 & 1.36 & - & 0.023 & 0.021 & 0.015 & 1.4 \\\cline{2-13}
        &Resnet101&44M&0.067 & 0.035 & 0.030 & 0.019 & 1.58 & - & 0.040 & 0.037 & 0.029 & 1.28 \\ \cline{2-13}
        &Resnet152&60M& 0.090 & 0.053 & 0.042 & 0.028 & 1.50 & - & 0.057 & 0.046 & 0.036 & 1.28 \\ \cline{2-13}
        &EfficientNet B0 &5M&0.063 & 0.025 & 0.020 & 0.010 & 2.00 & - & 0.029 & 0.025 & 0.012 & 2.08 \\ \cline{2-13}
        &EfficientNet B1 &7M&0.092 & 0.034 & 0.029 & 0.014 & 2.07 & - & 0.042 & 0.034 & 0.018 & 1.89 \\ \cline{2-13}
        &Densenet121&8M&0.054 & 0.046 & 0.035 & 0.021 & 1.67 & - & 0.058 & 0.042 & 0.025 & 1.68 \\ \cline{2-13}
        &Densenet169&14M&0.075 & 0.068 & 0.048 & 0.029 & 1.66 & - & 0.075 & 0.057 & 0.033 & 1.72\\\cline{2-13}
        &Densenet201&17M&0.089 & 0.081 & 0.057 & 0.035 & 1.63 & - & 0.086 & 0.066 & 0.039 & 1.69\\ \bottomrule \multicolumn{1}{r}{}&\multicolumn{6}{r}{arithmetic average speedup factor from fastest alternative:}&\multicolumn{1}{c}{{\bf 1.92}}&\multicolumn{4}{r}{}&\multicolumn{1}{c}{{\bf 1.55}} \\ 
        \multicolumn{1}{r}{}&\multicolumn{6}{r}{geometric average speedup from fastest alternative:}&\multicolumn{1}{c}{{\bf 1.80}}&\multicolumn{4}{r}{}&\multicolumn{1}{c}{{\bf 1.47}} \\\bottomrule
    \end{tabular}
    \label{tab:timesW}
\end{table*}

As clarified by the above implementation walk-through, our method offers two great advantages. First, the details and peculiarities of the considered DN do not impact nor complicate the implementation of our method. All that needs to be done is to replace the nonlinearity functions from the model declaration with the provided ones, everything else left untouched. Second, our method can be implemented in languages/libraries that are not equipped with automatic differentiation. This is an important benefit as it allows deployment in non deep learning focused platforms. Furthermore, specifically designed chips \cite{ando2017brein} can employ our solution without additional difficulties than when implementing a standard DN forward pass. The next section provides various experiments comparing the speed of the method compared to alternative solutions.

\subsection{Speed Validation Across Architectures and Hardware}
\label{sec:validation}

\begin{figure*}[t!]
    \centering
    \begin{minipage}{0.32\linewidth}
    \includegraphics[width=\linewidth]{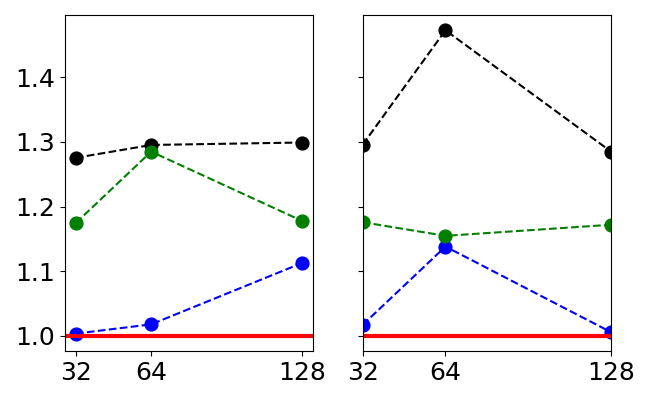}
    \end{minipage}
    \begin{minipage}{0.67\linewidth}
    \caption{\small Depiction of the average slow-down factor (over $1000$ runs) induced by employing the proposed custom nonlinearities from Algo.~\ref{algo:implementation} instead of the default ones for the activation in {\bf black}, and max-pooling in {\bf green} with $32$-channel inputs, varying spatial shape ({\bf x-axis}) and with mini-batch size of $10$ and $80$ ({\bf left,right}). The {\bf red} line is at $y=1$. We also depict the case of a Conv2D layer in {\bf blue}, in this case the line depicts the slow-down incurred by applying the Conv2D layer on an input with mini-batch size multiplied by $3$. We see that forwarding through those operators is at most $50\%$ slower than using the default nonlinearity/Conv2D with  mini-batch size divided by $3$, meaning that one JVP evaluation takes no more than $50\%$ more time than a direct input forward propagation in $f$.}
    \label{fig:speedups}
    \end{minipage}
\end{figure*}

We first propose in Fig.~\ref{fig:speedups} to depict the slow down that is incurred by using the custom nonlinearities from Algo.~\ref{algo:implementation} with input $X$ that concatenates $\bx,\bu,\mathbf{0}$ instead of the default nonlinearities with input $\bx$ for different input shapes and mini-batch sizes. As discussed in Sec.~\ref{sec:implementation} despite the need to clone the nonlinearity matrices and to process three times more inputs, we observe that the slow-down factor remains around $1.5$, at most, thanks to the highly parallel platform offered by GPUs. This serves as a good indicator to practitioners that the computation time of the JVP evaluation proposed in this paper will generally not exceed $1.5 \times$ the computation time needed to perform a forward pass of $\bx$ using the original DN that employs the default nonlinearities. We also propose in Table~\ref{tab:timesX} and Table~\ref{tab:timesW} the computation times to evaluate the JVP of multiple DN architectures taken with respect to their input and layer weights respectively. We employ various architectures and also compare different GPUs performances (in all case using a single GPU). We observe that our solution is the fastest with speed-up factors varying based on the considered architecture, ranging from $~1.1$ to $~7$. In addition of accelerating JVP evaluations, our solution allows JVP evaluation for models like RNNs where the alternative methods fail to provide any result from Out Of Memory failures most likely due to the automatic differentiation method overhead induced from the recurrence mechanism.

\section{Fast Jacobian-Vector Product Extensions}
\label{sec:extension}

\vspace{-0.1cm}

We now demonstrate how (fast) JVPs can be used to compute some important quantities that have been shown to characterize DNs e.g. the largest eigenvalues and eigenvectors of $\bA_{\omega}$, the per-region slope matrix that produces the DN output (recall (\ref{eq:DNmapping})). 
Recall that for large models, extracting $\bA_{\omega(\bx)}$ entirely, and then performing any sort of analysis is not feasible ($196608$ by $196608$ matrix for datasets a la ImageNet \cite{deng2009imagenet}). It is thus crucial to evaluate the desired quantities only through JVP calls.
\vspace{-0.3cm}
\subsection{Deep Network Spectroscopy}

\begin{algorithm}[t!]
\caption{\small Top-$k$ spectral decomposition of a square slope matrix $\bA_{\omega} \in \mathbb{R}^{D\times D}$. For the case of rectangular matrix see Algo.~\ref{algo:SVD} in the Appendix.}
\label{algo:eigen}
\begin{algorithmic}
\Procedure{}{$k \in \{1,\dots, K\},\bx \in \omega, tol >0$}
    \State randomly initialize $\bV=[\bv_1,\dots,\bv_k], \bv_i \in\mathbb{R}^{D}$
    \State $\bC\leftarrow \left[{\rm Rop}(f,\bx,\bv_1),\dots,{\rm Rop}(f,\bx,\bv_k)\right] $
    \Repeat
        \State $\bQ,\bR \leftarrow \text{QRDecomposition}(\bC)$
        \State $\bV \leftarrow \bQ_{.,1:k},\;\;\;\bSigma \leftarrow \bR_{1:k,.}$
        \State $\bC\leftarrow \left[{\rm Rop}(f,\bx,\bv_1),\dots,{\rm Rop}(f,\bx,\bv_k)\right] $
    \Until{$\|\bC-\bV\bSigma\| \leq tol$}\\
    \textbf{returns}: top-k eigenvalues $(\bSigma)$ and eigenvectors $(\bV)$
\EndProcedure
\end{algorithmic}
\end{algorithm}

\begin{figure}[t!]
    \centering
    \begin{minipage}{0.015\linewidth}
    \rotatebox{90}{\tiny $\log(\|\bC-\bV\bSigma\|)$}
    \end{minipage}
    \begin{minipage}{0.46\linewidth}
    \centering
    \includegraphics[width=1\linewidth]{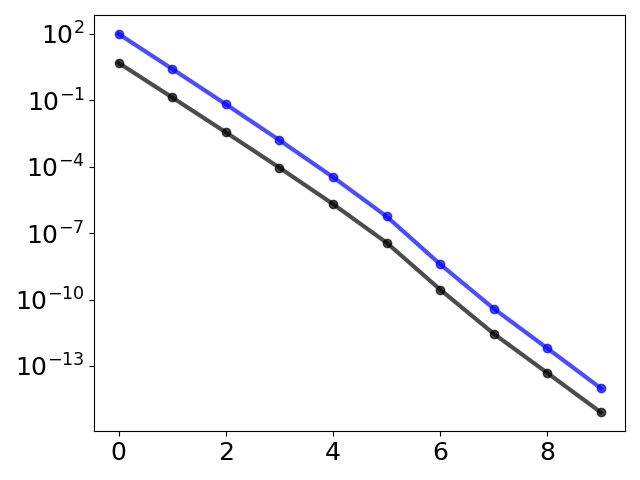}\\
    
    iterations
    \end{minipage}
    \begin{minipage}{0.022\linewidth}
    \rotatebox{90}{\tiny $\|\bA_{\omega(\bx)}\bz\|_2^2$}
    \end{minipage}
    \begin{minipage}{0.46\linewidth}
    \centering
    \includegraphics[width=1\linewidth]{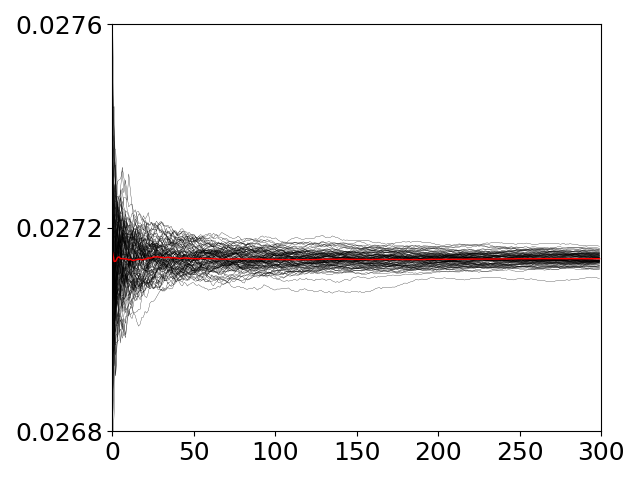}\\
    
    iterations
    \end{minipage}
    \vspace{-0.3cm}
    \caption{\small {\bf left}: depiction of the top-k eigenvectors/eigenvalues computation with a UNet based Algo.~\ref{algo:eigen} for $k=1$ and with $100$ in {\bf black} and for $k=50$ in {\bf blue}, we observe that convergence is obtained after only a few iterations, the $10$ iterations done for $100$ and $10$ samples simultaneously respectively for $k=1,50$ takes $3.12s$ and $16s$. {\bf Right}: different trajectories of Monte-Carlo estimation of (\ref{eq:Ex}) to estimate the Frobenius norm of $\bA_{\omega(\bx)}$ with a Unet and given a random sample $\bx$, $100$ MC samples are obtained in $0.013s$ for a single datum.}
    \label{fig:extensions}
\end{figure}

First, we demonstrate how the DN closing recipe allows to compute the per-region affine parameters.
\begin{prop}
The $Rop$ allows efficient computation of the per-region slope and bias parameters from (\ref{eq:explicit}) given a sample $\bx \in \omega$ 

\begin{align*}
        \bA_{\omega(\bx)}\hspace{-0.1cm}=\hspace{-0.1cm}\left[{\rm Rop}(f,\bx,\be_1),{\tiny ...},{\rm Rop}(f,\bx,\be_D)\right]\hspace{-0.1cm},\bb_{\omega(\bx)}\hspace{-0.1cm}=\hspace{-0.1cm}f_{\bx}(\mathbf{0}),
\end{align*}

where $\be_d$ represents the element of the $\mathbb{R}^D$ canonical.
\end{prop}

\vspace{-0.1cm}

Another quantity that can easily be obtained from the DN cloning and Rop evaluation concerns the spectrum of the slope matrix $\bA_{\omega(\bx)}$ which has been theoretically studied for DNs in the context of data augmentation \cite{lejeune2019implicit} and dropout \cite{gal2016dropout}. It is possible to efficiently compute the top-k spectral decomposition of such matrices through iterative procedures as given in Algo.~\ref{algo:eigen} from \cite{bentbib2015block}, leveraging the JVP, with same asymptotic complexity than a simple forward propagation of an input through a DN $f$ (see Appendix~\ref{sec:complexity} for the proof). Illustrative application on a UNet is provided in Fig.~\ref{fig:extensions}.
This case demonstrates the importance and ability of such analysis to finally enable tractable computations of key quantities for DNs and should open the door to tremendous applications of the methods and principled empirical validation of various theoretical studies as well as open the door to novel interpretability and visualization techniques.

\subsection{Frobenius Norm Estimation}

We also demonstrate how the JVP provides an efficient solution to estimate the Frobenius norm of the per-region slope matrix. Such norm has been theoretically studied for DNs and was found to be an indicator of Deep Generative Networks training failures \cite{tanielian2020learning}, rapid evaluation of it would thus allow practitioners to detect such cases. It can be done as per \citep{gudmundsson1995small} with

\begin{align}
    \mathbb{E}[\|\bA_{\omega(\bx)}\bu\|_2^2]=\|\bA_{\omega(\bx)}\|^2_{F},\label{eq:Ex}
\end{align}
with $\bu$ a random unit sphere gaussian random variable \citep{devroye1986sample,calafiore1999radial}. Hence, (\ref{eq:Ex}) can easily be estimated via Monte-Carlo sampling by repeatedly performing JVPs, as we demonstrate in Fig.~\ref{fig:extensions}. While the above illustrates a particular case, the same recipe can be employed to estimate ${\rm trace}(\bA_{\omega(\bx)})$ and $\log \det (I+\bA_{\omega(\bx)})$
for Hermitian positive semi-definite matrices \citep{saibaba2017randomized}, or to estimate $\|\bA_{\omega(\bx)}^{-1}\|_{F}$ \citep{kenney1998statistical}. For in-depth study of such  numerical methods that are built-upon JVPs, we refer the reader to \citet{trefethen1997numerical,doi:10.1080/00029890.1998.12004985,meyer2000matrix}.

\vspace{-0.3cm}
\section{Conclusions}

\vspace{-0.2cm}
Automatic differentiation and backpropagation are behind most if not all recent successes in  deep learning,  yielding not only state-of-the-art results but also fueling the discoveries of new architectures, normalizations, and the likes. This first wave of research almost exclusively relied on backward differentiation (gradient computations) but recently, many techniques further pushing DN performances have been relying on Jacobian-vector products (JVPs). Current JVPs evaluations rely on automatic differentiation making them slower and less memory efficient than direct evaluation which in turn requires to be implemented by hand based on the architecture employed. We demonstrated in this paper that current DNs, which employ nonlinearities such as (leaky-)ReLU and max-pooling, can have their JVPs evaluated directly without the need for automatic differentiation and without the need to hard-code each architecture only by changing a few lines of code in their implementation. We validated our proposed method against alternative implementations and demonstrated significant speed-ups of a factor of almost $2$ on average over $13$ different architectures and hardware. We also demonstrated that fast JVPs will help in the development of novel analysis and theoretical techniques for CPA DNs that require evaluations of quantities such as eigenvectors/eigenvalues of the per-region slope matrices that benefit from faster JVP evaluations.

\section{Acknowledgments}
This work was supported by NSF grants CCF-1911094, IIS-1838177, and IIS-1730574; ONR grants N00014-18-12571, N00014-20-1-2787, and N00014-20-1-2534; AFOSR grant FA9550-18-1-0478; and a Vannevar Bush Faculty Fellowship, ONR grant N00014-18-1-2047.

\bibliography{iclr2021_conference}
\bibliographystyle{icml2021}

\appendix

\section{Details on Jacobian Vector Products}
\label{sec:detailsJVP}

The differential allows to express the local growth of each function's output dimension as a linear term w.r.t the growth variable on each input dimension plus a small error term. This information is formatted into a Jacobian matrix which has the form
\begin{align*}
    \J_f(\bx) = \begin{pmatrix}
    \frac{f_1}{\partial x_1}(\bx)&\dots&\frac{f_1}{\partial x_D}(\bx)\\
    \vdots&\ddots&\vdots\\
    \frac{f_K}{\partial x_1}(\bx)&\dots&\frac{f_K}{\partial x_D}(\bx)
    \end{pmatrix},
\end{align*}
where $f_i$ is the $i^{\rm th}$ output dimension of the operator $f$. Note that one can also express the Jacobian matrix in term of the gradient operator $\nabla$ taken for each output dimension as
\begin{align*}
    \J_f(\bx) = \begin{pmatrix}
    \nabla_{f_1}(\bx)^T\\
    \vdots\\
    \nabla_{f_K}(\bx)^T
    \end{pmatrix}.
\end{align*}
One has the following property when assuming that $f$ is differentiable in $\bx \in U$ where $U$ is an open subspace of $\mathbb{R}^D$
$$f(\bx+\bu)=f(\bx)+ \J_{f}(\bx)\bu + o(\|\bu\|),$$
where it should be clear that the tangent space induced by the Jacobian matrix allows to locally describe the operator $f$ with approximation error increasing as one moves away from $\bx$, the point where the Jacobian matrix has been evaluated at.

\section{Proof of Rop Evaluation From Double Lop}
\label{sec:proofDJVP}

It is possible to compute the $Rop$ from two $Lop$ calls via
\begin{align*}
    {\rm Rop}(f,\bx,\bu)={\rm Lop}({\rm Lop}(f,\bx,\bv)^T,\bv,\bu)^T,
\end{align*}
which we now prove.
First, we have that 
\begin{align*}
    {\rm Lop}(f,\bx,\bv)^T=&\left(\bv^T\J_{f}(\bx)\right)^T\\
    =&\J_{f}(\bx)^T\bv,
\end{align*}
denote the following linear mapping in $\bv$ as $g(\bv,\bx)=\J_{f}(\bx)^T\bv$, we have
\begin{align*}
{\rm Lop}({\rm Lop}(f,\bx,\bv)^T,\bv,\bu)^T=&{\rm Lop}(g,\bv,\bu)^T\\
=&\left(\bu^T\J_{f}(\bx)^T\right)^T\\
=&\J_{f}(\bx)\bu\\
=&{\rm Rop}(f,\bx,\bu),
\end{align*}
which completes the proof.

\section{Max-Pooling Nonlinearity Matrix}
\label{appendix:maxpooling}

We saw in (\ref{eq:Q}) how the nonlinearity matrix for activation operators are square, diagonal, and with entries based on the employed activations e.g. $0,1$ for ReLU and $-1,1$ for absolute value. This is true as the activation operator maintains the same dimensions from its input to its output. A max-pooling operator however reduces the dimensionality of its input based on some policy, for concreteness say by performing a spatial pooling over $2\times2$ non-overlapping regions of its input. Hence the $\bQ^{\ell}$ matrix of the max-pooling operator will be rectangular, and in this particular case, with number of rows equal to the number of columns (input dimension) divided by $4$, the size of the pooling regions. Since we consider flattened versions of the input tensors, let first recall that when employing such a pooling with a tensor, the output at spatial position and channel  $(i,j, c)$ in the output performs pooling over the indices $(i,j,c),(i+1,j,c),(i,j+1,c),(i+1,j+1,c)$. This translates into the flattened vectors into indices $((i-1)*W+j)*C+c,((i-1)*W+j+1)*C+c,(i*W+j)*C+c,(i*W+j+1)*C+c$ for the input vector and index $((i-1)*W/2+j)*C+c$ for the output vector. We also have in this case that the input tensor is of shape $(H,W,C)$ and output tensor is $(H/2,W/2,C)$.
As a result, the rectangular matrix $\bQ^{\ell}$ will be filled with $0$, with a single $1$ per row, the position one this $1$ in each row can be in any index from $((i-1)*W+j)*C+c,((i-1)*W+j+1)*C+c,(i*W+j)*C+c,(i*W+j+1)*C+c$ based on the argmax position from the input values at those indices.

\section{Implementation Details}
\label{sec:details}

\subsection{Software}

We employed the latest stable version of Tensorflow, at the time of our study this was version 2.4.0, for all the additional details and versions of supporting libraries like Numpy we refer the reader to our GitHub page.

\subsection{Hardware}

{\bf TITAN X (Pascal)}: The TITAN X Pascal is an enthusiast-class graphics card by NVIDIA, launched in August 2016. Built on the 16 nm process, and based on the GP102 graphics processor, in its GP102-400-A1 variant, the card supports DirectX 12. This ensures that all modern games will run on TITAN X Pascal. The GP102 graphics processor is a large chip with a die area of 471 mm$^2$ and 11,800 million transistors. Unlike the fully unlocked TITAN Xp, which uses the same GPU but has all 3840 shaders enabled, NVIDIA has disabled some shading units on the TITAN X Pascal to reach the product's target shader count. It features 3584 shading units, 224 texture mapping units, and 96 ROPs. NVIDIA has paired 12 GB GDDR5X memory with the TITAN X Pascal, which are connected using a 384-bit memory interface. The GPU is operating at a frequency of 1417 MHz, which can be boosted up to 1531 MHz, memory is running at 1251 MHz (10 Gbps effective).
Being a dual-slot card, the NVIDIA TITAN X Pascal draws power from 1x 6-pin + 1x 8-pin power connector, with power draw rated at 250 W maximum. Display outputs include: 1x DVI, 1x HDMI, 3x DisplayPort. TITAN X Pascal is connected to the rest of the system using a PCI-Express 3.0 x16 interface. The card's dimensions are 267 mm x 112 mm x 40 mm, and it features a dual-slot cooling solution. Its price at launch was 1199 US Dollars.

{\bf GeForce GTX 1080 Ti}: The GeForce GTX 1080 Ti is an enthusiast-class graphics card by NVIDIA, launched in March 2017. Built on the 16 nm process, and based on the GP102 graphics processor, in its GP102-350-K1-A1 variant, the card supports DirectX 12. This ensures that all modern games will run on GeForce GTX 1080 Ti. The GP102 graphics processor is a large chip with a die area of 471 mm$^2$ and 11,800 million transistors. Unlike the fully unlocked TITAN Xp, which uses the same GPU but has all 3840 shaders enabled, NVIDIA has disabled some shading units on the GeForce GTX 1080 Ti to reach the product's target shader count. It features 3584 shading units, 224 texture mapping units, and 88 ROPs. NVIDIA has paired 11 GB GDDR5X memory with the GeForce GTX 1080 Ti, which are connected using a 352-bit memory interface. The GPU is operating at a frequency of 1481 MHz, which can be boosted up to 1582 MHz, memory is running at 1376 MHz (11 Gbps effective).
Being a dual-slot card, the NVIDIA GeForce GTX 1080 Ti draws power from 1x 6-pin + 1x 8-pin power connector, with power draw rated at 250 W maximum. Display outputs include: 1x HDMI, 3x DisplayPort. GeForce GTX 1080 Ti is connected to the rest of the system using a PCI-Express 3.0 x16 interface. The card's dimensions are 267 mm x 112 mm x 40 mm, and it features a dual-slot cooling solution. Its price at launch was 699 US Dollars.

{\bf Quadro RTX 8000}: The Quadro RTX 8000 is an enthusiast-class professional graphics card by NVIDIA, launched in August 2018. Built on the 12 nm process, and based on the TU102 graphics processor, in its TU102-875-A1 variant, the card supports DirectX 12 Ultimate. The TU102 graphics processor is a large chip with a die area of 754 mm$^2$ and 18,600 million transistors. It features 4608 shading units, 288 texture mapping units, and 96 ROPs. Also included are 576 tensor cores which help improve the speed of machine learning applications. The card also has 72 raytracing acceleration cores. NVIDIA has paired 48 GB GDDR6 memory with the Quadro RTX 8000, which are connected using a 384-bit memory interface. The GPU is operating at a frequency of 1395 MHz, which can be boosted up to 1770 MHz, memory is running at 1750 MHz (14 Gbps effective).
Being a dual-slot card, the NVIDIA Quadro RTX 8000 draws power from 1x 6-pin + 1x 8-pin power connector, with power draw rated at 260 W maximum. Display outputs include: 4x DisplayPort, 1x USB Type-C. Quadro RTX 8000 is connected to the rest of the system using a PCI-Express 3.0 x16 interface. The card measures 267 mm in length, 111 mm in width, and features a dual-slot cooling solution. Its price at launch was 9999 US Dollars.

{\bf CPU:} Architecture:        x86\_64,
CPU op-mode(s):      32-bit, 64-bit,
Byte Order:          Little Endian,
CPU(s):              96,
On-line CPU(s) list: 0-95,
Thread(s) per core:  2,
Core(s) per socket:  24,
Socket(s):           2,
NUMA node(s):        2,
Vendor ID:           GenuineIntel,
CPU family:          6,
Model:               85,
Model name:          Intel(R) Xeon(R) Gold 5220R CPU @ 2.20GHz,
Stepping:            7,
CPU MHz:             2200.000,
CPU max MHz:         2201.0000,
CPU min MHz:         1000.0000,
BogoMIPS:            4400.00,
Virtualization:      VT-x,
L1d cache:           32K,
L1i cache:           32K,
L2 cache:            1024K,
L3 cache:            36608K,
NUMA node0 CPU(s):   0-23,48-71,
NUMA node1 CPU(s):   24-47,72-95.

\section{Layer JVP Implementation}
\label{sec:layer_implementation}
For the layer case, that is, taking the JVP of the DN output at $\bx$ with respect to a layer weight $\bW^{\ell}$, the implementation is similar except that all the first layers up to and including $\ell-1$ remain with their default nonlinearities. All the layers starting and including $\ell$ will employ the custom nonlinearities that we provided. At layer $\ell$, the user will take the layer input and generate the $X$ tensor as follows for the case of a dense layer, we denote by $\bW$, $\bU$ the considered layer weight and the direction to produce the JVP from
\begin{python}
x = previous_layer(x) # x is now the 
                      # current layer input
hx = tf.matmul(x,W)+b # this is the usual 
                      # pre-activation
hu = tf.matmul(x,U-W)+b # same but using the 
                      # given direction
h0 = b
X = tf.concat([hx,hu], 0)
output = clone_act(X,0.1)
\end{python}
the same goes for the case of a convolutional layer as in
\begin{python}
x = previous_layer(x)
hx = tf.nn.conv2d(x,W,strides,padding)+b
hu = tf.nn.conv2d(x,U-W,strides,padding)+b
h0 = b
X = tf.concat([hx,hu], 0)
output = clone_act(X,0.1)
\end{python}
if one wanted to use batch-normalization, add a \pyth{clone_bn} operation just before the activation, taking X as input, the output of this cloned BN would then be fed into the \pyth{clone_act} operation.

\section{Codes for Additional Nonlinearities}
\label{sec:codes}

In the case of batch-normalization layer, the input output mapping is not a CPA and thus we need to explicitly compute by hand the Jacobian of the layer and implement the forward pass. This is only needed if one wants to compute the JVP during training, otherwise this can be omitted since during testing the batch-normalization is simply an affine transformation of the input.

\begin{algorithm}[h]
\caption{{\small Custom nonlinearity implementations to be used jointly with the DN input that concatenates $\bx,\bu,\mathbf{0}$ (recall (\ref{eq:JVPclone}) and (\ref{eq:JVPclone2})). Those implementations are drop-in replacements of the usual ones.}}
\label{algo:implementation}
\begin{python}
def clone_maxpool(X, ksize, strides, pad):
  K, N = len(X.shape)-1, X.shape[0]
  x_only = X[: N//2]
  argmax = tf.nn.max_pool_with_argmax(
            x_only, ksize, strides, pad,
            include_batch_in_index=True)[1]
  n_vals = x_only.shape.num_elements()
  shifts = tf.repeat(tf.range(2)*n_vals, 
                     [N//2])
  shifts = tf.reshape(shifts, [-1]+[1]*K)
  flat = tf.reshape(X, (-1,))
  t_argmax = tf.tile(argmax, [2]+[1]*K)
  return tf.gather(flat, 
          shifts+tf.cast(t_argmax,'int32'))
          
def clone_dropout(X, rate, training=None):
  if ! training:
    return X
  keep_prob, K = 1 - rate, len(X.shape)-1
  n_shape = (X.shape[0]//2,) + X.shape[1:]
  mask = tf.random.uniform(n_shape) > rate
  mask = tf.tile(mask, [2] + [1] * K)
  return X * mask / keep_prob
\end{python}
\end{algorithm}

\section{Proof}

\subsection{Proof of Theorem~\ref{thm:JVP}}
\label{proof:JVP}
\begin{proof}
The proof for the DN input JVP case is trivial and given in (\ref{eq:JVPclone}). Now when considering the layer weight case, the only added complication comes from the fact that the parameter of interest is a matrix (as opposed to a vector). However, we will simply demonstrate that a layer output can be produced by considering the flattened version of the weight matrix, in which case the JVP is taken w.r.t. a vector.

Denote a vector $\bw$ encoding the flattened representation of $\bW^{\ell}$, the DN mapping based on an input $\bx$ but now taking as input $\bw$ is defined as
\begin{align}
\tilde{f}_{\bx}(\bw)\hspace{-0.09cm}=\hspace{-0.09cm}\left(f^{L}\circ \dots \circ f^{\ell+1}\right)\hspace{-0.09cm}\left(\bh^{\ell}(\bZ\bw+\bb^{\ell})\right),
\end{align}
with $\bZ$ a $D^{\ell}$ by $D^{\ell}\times D^{\ell-1}$ matrix stacking repeatedly $\bz^{\ell-1}(\bx)$ as $\bZ=[\bz^{\ell-1}(\bx),\dots,\bz^{\ell-1}(\bx)]^T$. 
First, notice that layers $1,\dots,\ell-1$ do not depend on $\bW^{\ell}$. 
From this, it is possible to apply the same strategy as when taking the JVP with the DN since by considering $\tilde{f}$ the DN input is now the layer weight of interest as in
$$
{\rm Rop}(f,\bx,\bu)=\tilde{f}_{\bx}(\bu)-\tilde{f}_{\bx}(\mathbf{0}).
$$
However, one can further simplify the above such that in practice there is no need to produce the $\bZ$ matrix explicitly. Notice that for $\tilde{f}_{\bx}(\bu)$, the layer will compute $\bZ\bu+\bb^{\ell}$. This quantity, is nothing else that $\bU\bz^{\ell-1}(\bx)+\bb^{\ell}$ where $\bU$ is now the matrix form of $\bu$ and the matrix $\bZ$ is put back in vector  form (and without the duplication) as $\bz^{\ell}(\bx)$. We thus directly obtain that in practice, computing the above is equivalent to computing the layer output but replacing the layer weights by $\bU$ and $\mathbf{0}$ matrices. For the convolutional layer case, the exact same process is applied but now noticing that $\bW^{\ell}$ can be written as the product between the (small) weight vectors and a matrix producing the circulant block circulant matrix of convolution.
\end{proof}

\section{Computational Complexity of Spectral Decomposition}
\label{sec:complexity}

The QR decomposition \citep{strang2019linear} using Householder transformations is done with $2dn^2-2n^3/3$ flops \cite{golub1996matrix} for a $d\times n$ matrix. Hence the top-$k$ SVD is obtained by repeating until reaching the stopping criteria $k$ $Rop$ calls and $2dk^2-2k^3/3$ additional flops. Considering a fixed number of iteration of the inner-loop we obtain that the asymptotic time-complexity of Algo.~\ref{algo:eigen} is the same as the asymptotic time-complexity of performing a forward pass into the considered DN.

\section{Additional Results}

\begin{table*}[t!]
    \centering
    \setlength\tabcolsep{3pt} 
    \renewcommand{\arraystretch}{0.9}
    \caption{Reprise of Table~\ref{tab:timesX} but now considering $Rop(f(,\bx,\bu)$ for GeForce GTX 1080Ti.}
    \begin{tabular}{c|l|l||rrrr|r|}
        \cline{4-8}
        \multicolumn{3}{r|}{}&\multicolumn{5}{c|}{$\bx \in \mathbb{R}^{100\times 100 \times 3},\by\in\mathbb{R}^{20}$}\\ \cline{2-8}
        \multicolumn{1}{r|}{}&Model & \# params. & \makecell{batch-\\jacobian} & jvp & \makecell{double \\ vjp} & \makecell{clone \\ (ours)} & \makecell{ speedup \\ factor}\\ \toprule
        \multirow{13}{*}{\rotatebox{90}{GeForce GTX 1080Ti}}
        &RNN&5M&- & - & 0.166 & 0.034 & 4.88 \\ \cline{2-8}
        &UNet &49M&- & - & - & - & - \\ \cline{2-8}
         &VGG16 &138M& 0.060 & 0.028 & 0.024 & 0.017 & 1.41 \\
        \cline{2-8}
        &VGG19&143M&0.075 & 0.039 & 0.031 & 0.021 & 1.47 \\ \cline{2-8}
        &Inception V3&23M&0.041 & 0.022 & 0.017 & 0.011 & 1.54 \\\cline{2-8}
        &Resnet50&25M&0.041 & 0.014 & 0.012 & 0.011 & 1.09  \\\cline{2-8}
        &Resnet101&44M&0.058 & 0.021 & 0.017 & 0.017 & 1  \\ \cline{2-8}
        &Resnet152&60M&  0.082 & 0.03 & 0.025 & 0.025 & 1  \\ \cline{2-8}
        &EfficientNet B0 &5M& - & - & - & 0.007 & $\infty$ \\ \cline{2-8}
        &EfficientNet B1 &7M&- & - & - & 0.009 & $\infty$  \\ \cline{2-8}
        &Densenet121&8M&0.037 & 0.026 & 0.019 & 0.013 & 1.46\\ \cline{2-8}
        &Densenet169&14M&0.045 & 0.036 & 0.026 & 0.018 & 1.44\\\cline{2-8}
        &Densenet201&17M&0.061 & 0.046 & 0.033 & 0.018 & 1.83\\ \bottomrule
    \end{tabular}
    \label{tab:timesWcpu}
\end{table*}

\begin{table*}[t!]
    \centering
    \setlength\tabcolsep{3pt} 
    \renewcommand{\arraystretch}{0.9}
    \caption{Reprise of Table~\ref{tab:timesW} but now considering $Rop(f(\bx),\bW,\bU)$ for GeForce GTX 1080Ti.}
     \begin{tabular}{c|l|l||rrrr|r|}
        \cline{4-8}
        \multicolumn{3}{r|}{}&\multicolumn{5}{c|}{$\bx \in \mathbb{R}^{100\times 100 \times 3},\by\in\mathbb{R}^{20}$}\\ \cline{2-8}
        \multicolumn{1}{r|}{}&Model & \# params. & \makecell{batch-\\jacobian} & jvp & \makecell{double \\ vjp} & \makecell{clone \\ (ours)} & \makecell{ speedup \\ factor}\\ \toprule
        \multirow{13}{*}{\rotatebox{90}{GeForce GTX 1080Ti}}
        &RNN&5M&- & - & 0.151 & 0.024 & 6.29167 \\ \cline{2-8}
        &UNet &49M&- & - & - & - & - \\ \cline{2-8}
         &VGG16 &138M& 0.060 & 0.028 & 0.024 & 0.017 & 1.41 \\
        \cline{2-8}
        &VGG19&143M&0.072 & 0.036 & 0.031 & 0.022 & 1.40 \\ \cline{2-8}
        &Inception V3&23M&0.039 & 0.021 & 0.016 & 0.011 & 1.45 \\\cline{2-8}
        &Resnet50&25M&0.039 & 0.013 & 0.011 & 0.011 & 1  \\\cline{2-8}
        &Resnet101&44M&0.058 & 0.023 & 0.017 & 0.017 & 1  \\ \cline{2-8}
        &Resnet152&60M& 0.080 & 0.029 & 0.027 & 0.025 & 1.08  \\ \cline{2-8}
        &EfficientNet B0 &5M& - & - & - & 0.007 & $\infty$ \\ \cline{2-8}
        &EfficientNet B1 &7M&- & - & - & 0.009 & $\infty$  \\ \cline{2-8}
        &Densenet121&8M&0.033 & 0.025 & 0.020 & 0.011 & 1.81\\ \cline{2-8}
        &Densenet169&14M&0.050 & 0.036 & 0.026 & 0.015 & 1.73\\\cline{2-8}
        &Densenet201&17M&0.061 & 0.044 & 0.030 & 0.018 & 1.67\\ \bottomrule
    \end{tabular}
    \label{tab:timesXcpu}
\end{table*}

\begin{algorithm}[h]
\caption{Top-$k$ SVD \cite{golub1965calculating}  of the slope matrix $\bA_{\omega(\bx)} \in \mathbb{R}^{K\times D}$.}
\label{algo:SVD}
\begin{algorithmic}
\Procedure{BlockSVD}{$k,\bx \in \omega$}
    \State randomly initialize $\bU =[\bu_1,\dots,\bu_{k}]\in\mathbb{R}^{D \times k}$
    \State randomly initialize $\bV=[\bv_1,\dots,\bv_{k}] \in\mathbb{R}^{D \times k}$
    \State randomly initialize $\bSigma \in\mathbb{R}^{k \times k}$
    \State $\bC\leftarrow \left[{\rm Rop}(f,\bx,\bv_1),\dots,{\rm Rop}(f,\bx,\bv_k)\right] $
    \While{$\|\bC-\bU\bSigma\| > tol$}
        \State $\bQ,\bR \leftarrow \text{QRDecomposition}(\bC)$
        \State $\bU \leftarrow \left[\bQ\right]_{.,1:k}$\Comment{Update left singular vectors}
        \State $\bB\leftarrow \left[{\rm Lop}(f,\bx,\bu_1)^T,\dots,{\rm Lop}(f,\bx,\bu_k)^T\right]$
        \State $\bQ,\bR \leftarrow \text{QRDecomposition}(\bB)$
        \State $\bV \leftarrow \left[\bQ\right]_{.,1:k}$\Comment{Update right singular vectors}
        \State $\bSigma \leftarrow \left[\bR\right]_{1:k,1:k}$\Comment{Update singular values}
        \State $\bC\leftarrow \left[{\rm Rop}(f,\bx,\bv_1),\dots,{\rm Rop}(f,\bx,\bv_k)\right] $
    \EndWhile
    \textbf{returns}: top-k eigenvalues $(\bSigma)$ and eigenvectors $(\bV)$
\EndProcedure
\end{algorithmic}
\end{algorithm}

\begin{table*}[h]
    \centering
    \setlength\tabcolsep{6pt} 
    \renewcommand{\arraystretch}{1.1}
    \caption{Standard deviations of the computation times to perform ${\rm Rop}(f,\bx,\bu)$ over 1000 runs, - indicates Out Of Memory (OOM) failure}
    \begin{tabular}{|l|l||rrrr|rrrr|}
        \cline{3-10}
        \multicolumn{2}{r|}{}&\multicolumn{4}{c||}{$\bx \in \mathbb{R}^{100\times 100 \times 3},\by\in\mathbb{R}^{20}$}&\multicolumn{4}{c|}{$\bx \in \mathbb{R}^{400\times 400 \times 3},\by\in\mathbb{R}^{1000}$}\\ \hline
        Model & \# params. & \makecell{batch-\\jacobian} & jvp & \makecell{double \\ vjp} & \makecell{clone \\ (ours)} &  \makecell{batch-\\jacobian} & jvp & \makecell{double \\ vjp} & \makecell{clone \\ (ours)}\\ \toprule
        RNN &5M& 0.011 & - & 0.014 & 0.004 & - & - & 0.046 & 0.016 \\
        \hline
        UNet &49M& - & 0.004 & 0.003 & 0.002 & - & 0.006 & 0.004 & 0.005 \\
        \hline
         VGG16 &138M& 0.004 & 0.003 & 0.003 & 0.001 & - & 0.008 & 0.008 & 0.004 \\
        \hline
        VGG19&143M&0.005 & 0.003 & 0.003 & 0.002 & - & 0.009 & 0.008 & 0.005 \\ \hline
        Inception V3&23M&0.006 & 0.005 & 0.004 & 0.003 & - & 0.006 & 0.004 & 0.004 \\\hline
        Resnet50&25M&0.005 & 0.003 & 0.003 & 0.003 & - & 0.005 & 0.004 & 0.004 \\\hline
        Resnet101&44M&0.005 & 0.005 & 0.004 & 0.003 & - & 0.008 & 0.007 & 0.006 \\ \hline
        Resnet152&60M&0.010 & 0.006 & 0.004 & 0.004 & - & 0.010 & 0.006 & 0.007 \\ \hline
        EfficientNet B0 &5M&0.007 & 0.004 & 0.003 & 0.003 & - & 0.006 & 0.004 & 0.003 \\ \hline
        EfficientNet B1 &7M&0.007 & 0.005 & 0.004 & 0.003 & - & 0.007 & 0.006 & 0.004 \\ \hline
        Densenet121&8M& 0.007 & 0.005 & 0.005 & 0.003 & - & 0.008 & 0.006 & 0.005 \\ \hline
        Densenet169&14M&0.012 & 0.008 & 0.005 & 0.005 & - & 0.008 & 0.008 & 0.006\\\hline
        Densenet202&17M&0.013 & 0.011 & 0.007 & 0.004 & - & 0.008 & 0.012 & 0.007\\ \midrule
        RNN &5M& - & - & 0.022 & 0.007 & - & - & - & 0.017 \\
        \hline
        UNet &49M&- & 0.004 & 0.003 & 0.002 & - & 0.01 & 0.006 & 0.007 \\
        \hline
         VGG16 &138M& 0.007 & 0.002 & 0.001 & 0.001 & - & 0.009 & 0.005 & 0.005 \\
        \hline
        VGG19&143M&0.004 & 0.003 & 0.002 & 0.001 & - & 0.010 & 0.005 & 0.007\\ \hline
        Inception V3&23M&0.007 & 0.007 & 0.006 & 0.003 & - & 0.006 & 0.005 & 0.004 \\\hline
        Resnet50&25M&0.005 & 0.003 & 0.003 & 0.002 & - & 0.002 & 0.002 & 0.004 \\\hline
        Resnet101&44M&0.006 & 0.004 & 0.005 & 0.004 & - & 0.004 & 0.004 & 0.007 \\ \hline
        Resnet152&60M&0.006 & 0.009 & 0.007 & 0.005 & - & 0.003 & 0.004 & 0.004\\ \hline
        EfficientNet B0 &5M&0.011 & 0.004 & 0.003 & 0.001 & - & 0.002 & 0.001 & 0.001 \\ \hline
        EfficientNet B1 &7M&0.014 & 0.006 & 0.005 & 0.002 & - & 0.007 & 0.001 & 0.002 \\ \hline
        Densenet121&8M& 0.011 & 0.009 & 0.006 & 0.004 & - & 0.008 & 0.004 & 0.002 \\ \hline
        Densenet169&14M&0.015 & 0.011 & 0.008 & 0.005 & - & 0.006 & 0.004 & 0.004\\\hline
        Densenet202&17M&0.017 & 0.012 & 0.011 & 0.005 & - & 0.007 & 0.004 & 0.004\\
        \bottomrule
    \end{tabular}
    \label{tab:std_x}
\end{table*}

\begin{table*}[h]
    \centering
    \setlength\tabcolsep{6pt} 
    \renewcommand{\arraystretch}{1.1}
    \caption{Standard deviations of the computation times to perform ${\rm Rop}(f(\bx),\bW^{1},\bU)$ over 1000 runs, - indicates Out Of Memory (OOM) failure}
    \begin{tabular}{|l|l||rrrr|rrrr|}
        \cline{3-10}
        \multicolumn{2}{r|}{}&\multicolumn{4}{c||}{$\bx \in \mathbb{R}^{100\times 100 \times 3},\by\in\mathbb{R}^{20}$}&\multicolumn{4}{c|}{$\bx \in \mathbb{R}^{400\times 400 \times 3},\by\in\mathbb{R}^{1000}$}\\ \hline
        Model & \# params. & \makecell{batch-\\jacobian} & jvp & \makecell{double \\ vjp} & \makecell{clone \\ (ours)} &  \makecell{batch-\\jacobian} & jvp & \makecell{double \\ vjp} & \makecell{clone \\ (ours)}\\ \toprule
        RNN &5M& 0.013 & - & 0.013 & 0.006 & - & - & 0.042 & 0.018 \\
        \hline
        UNet &49M& - & 0.004 & 0.003 & 0.002 & - & 0.005 & 0.004 & 0.005 \\
        \hline
         VGG16 &138M& 0.004 & 0.003 & 0.003 & 0.001 & - & 0.008 & 0.008 & 0.003 \\
        \hline
        VGG19&143M&0.004 & 0.003 & 0.003 & 0.002 & - & 0.008 & 0.008 & 0.005 \\ \hline
        Inception V3&23M&0.006 & 0.004 & 0.004 & 0.004 & - & 0.007 & 0.004 & 0.004 \\\hline
        Resnet50&25M&0.003 & 0.003 & 0.003 & 0.003 & - & 0.004 & 0.004 & 0.003 \\\hline
        Resnet101&44M&0.007 & 0.005 & 0.004 & 0.003 & - & 0.007 & 0.004 & 0.004 \\ \hline
        Resnet152&60M&0.008 & 0.006 & 0.007 & 0.005 & - & 0.007 & 0.008 & 0.004 \\ \hline
        EfficientNet B0 &5M&0.004 & 0.003 & 0.003 & 0.003 & - & 0.006 & 0.004 & 0.003 \\ \hline
        EfficientNet B1 &7M&0.007 & 0.005 & 0.004 & 0.003 & - & 0.007 & 0.005 & 0.004 \\ \hline
        Densenet121&8M& 0.006 & 0.005 & 0.004 & 0.004 & - & 0.006 & 0.006 & 0.004 \\ \hline
        Densenet169&14M&0.012 & 0.006 & 0.006 & 0.004 & - & 0.014 & 0.007 & 0.006\\\hline
        Densenet202&17M&0.010 & 0.007 & 0.006 & 0.005 & - & 0.009 & 0.008 & 0.006\\ \midrule
        RNN &5M& - & - & 0.02 & 0.009 & - & - & - & 0.024 \\
        \hline
        UNet &49M& - & 0.004 & 0.002 & 0.002 & - & 0.009 & 0.003 & 0.007  \\
        \hline
         VGG16 &138M& 0.008 & 0.002 & 0.002 & 0.001 & - & 0.009 & 0.005 & 0.005 \\
        \hline
        VGG19&143M&0.004 & 0.003 & 0.002 & 0.001 & - & 0.009 & 0.006 & 0.007 \\ \hline
        Inception V3&23M&0.008 & 0.007 & 0.005 & 0.003 & - & 0.006 & 0.003 & 0.003 \\\hline
        Resnet50&25M&0.006 & 0.004 & 0.001 & 0.002 & - & 0.002 & 0.002 & 0.004 \\\hline
        Resnet101&44M&0.007 & 0.007 & 0.006 & 0.004 & - & 0.005 & 0.003 & 0.007  \\ \hline
        Resnet152&60M&0.007 & 0.009 & 0.008 & 0.006 & - & 0.007 & 0.007 & 0.005 \\ \hline
        EfficientNet B0 &5M&0.010 & 0.005 & 0.004 & 0.002 & - & 0.005 & 0.002 & 0.002 \\ \hline
        EfficientNet B1 &7M&0.029 & 0.007 & 0.005 & 0.003 & - & 0.009 & 0.001 & 0.003 \\ \hline
        Densenet121&8M& 0.011 & 0.010 & 0.007 & 0.004 & - & 0.009 & 0.004 & 0.003 \\ \hline
        Densenet169&14M&0.013 & 0.011 & 0.009 & 0.005 & - & 0.008 & 0.006 & 0.004\\\hline
        Densenet202&17M&0.014 & 0.013 & 0.010 & 0.005 & - & 0.01 & 0.003 & 0.002\\ \bottomrule
    \end{tabular}
    \label{tab:std_w}
\end{table*}

\end{document}